%% file: arxiv.tex
\documentclass{article}

\usepackage{arxiv}
\usepackage[labelfont=bf]{caption}
\usepackage[utf8]{inputenc} % allow utf-8 input
\usepackage[T1]{fontenc}    % use 8-bit T1 fonts
\usepackage{hyperref}       % hyperlinks
\usepackage{url}            % simple URL typesetting
\usepackage{booktabs}       % professional-quality tables
\usepackage{amsfonts}       % blackboard math symbols
\usepackage{nicefrac}       % compact symbols for 1/2, etc.
\usepackage{microtype}      % microtypography
\usepackage{lipsum}

\usepackage{units}
\usepackage{textcomp}
\usepackage{amsmath}
\usepackage{graphicx}
\usepackage{wasysym}
\usepackage{float}

\makeatletter
\usepackage{makeidx}
\makeatother

\AtBeginDocument{}

\begin{document}
\input{preamble.tex}

\input{arxivtitleauthors.tex}

\input{abstract.tex}

\input{introduction.tex}

\input{stateoftheart.tex}

\input{methodology.tex}

\input{experiments.tex}

\input{conclusion.tex}

\input{bibliography.tex}

\end{document}

%% file: preamble.tex
\renewcommand{\floatpagefraction}{.8}%

%% file: arxivtitleauthors.tex
\institute{Max Planck Institute for Intelligent Systems, Tübingen, Germany,\\
 \email{eric.price@tuebingen.mpg.de}\and Institute for Flight Mechanics and Controls, The Faculty of Aerospace Engineering and Geodesy, University of Stuttgart, Stuttgart, Germany.}

%\docdate{\today}
\docdate{2020-12-31}

\author{Eric Price\inst{1,2} \And Yu Tang Liu\inst{1,2} \And Michael J. Black\inst{1} \And Aamir Ahmad\inst{2,1} \thanks{We want to acknowledge the help of everyone in the Robot Perception Group at the Max Planck Institute for Intelligent Systems for their often spontaneous assistance with outdoor flight experiments. Special thanks goes to our past and current interns, most importantly to Pascal Goldschmid, for their tireless work on the robotic hardware and on board computers. Additional thanks to Melanie Feldhofer, for her assistance with flight permissions and insurance and Noa Price for help with manuscript editing.}}
%\author{
%  Eric Price \\
%  Max Planck Institute for Intelligent Systems\\
%  Tübingen, Germany \\
%  Institute for Flight Mechanics and Controls\\
%  \texttt{eric.price@tuebingen.mpg.de} \\
%  University of Stuttgart\\
%  Stuttgart, Germany \\
%  \And
% Yu–Tang Liu \\
%  Max Planck Institute for Intelligent Systems\\
%  Tübingen, Germany \\
%  Institute for Flight Mechanics and Controls\\
%  University of Stuttgart\\
%  Stuttgart, Germany \\
%  \And
% Michael J. Black \\
%  Max Planck Institute for Intelligent Systems\\
%  Tübingen, Germany \\
%  \And
% Aamir Ahmad \\
%  Max Planck Institute for Intelligent Systems\\
%  Tübingen, Germany \\
%  Institute for Flight Mechanics and Controls\\
%  University of Stuttgart\\
%  Stuttgart, Germany \\
%}

\title{Simulation and Control of Deformable Autonomous Airships in Turbulent Wind}\maketitle

\date{2020-12-28}

%% file: abstract.tex
\begin{abstract}
Fixed wing and multirotor UAVs are common in the field of robotics. Solutions for simulation and control of these vehicles are ubiquitous. This is not the case for airships, a simulation of which needs to address unique properties, i) dynamic deformation in response to aerodynamic and control forces, ii) high susceptibility to wind and turbulence at low airspeed, iii) high variability in airship designs regarding placement, direction and vectoring of thrusters and control surfaces. We present a flexible framework for modeling, simulation and control of airships, based on the Robot operating system (ROS), simulation environment (Gazebo) and commercial off the shelf (COTS) electronics, both of which are open source. Based on simulated wind and deformation, we predict substantial effects on controllability, verified in real world flight experiments. All our code is shared as open source, for the benefit of the community and to facilitate lighter-than-air vehicle (LTAV) research. \footnote{Source code: \href{https://github.com/robot-perception-group/airship_simulation}{https://github.com/robot-perception-group/airship\_{}{}simulation}}

\end{abstract}

%% file: introduction.tex
\begin{figure}[H]
\noindent \begin{centering}
{\small{}\includegraphics[width=1\columnwidth]{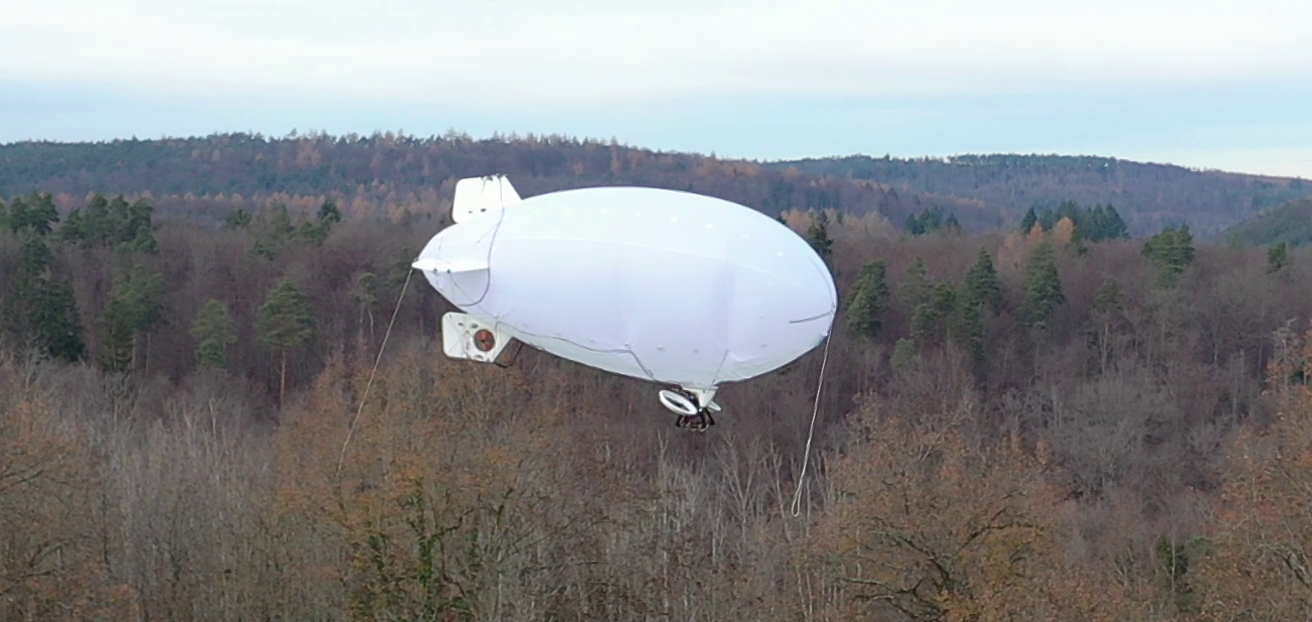}}{\small\par}
\par\end{centering}
{\small{}\caption{{\small{}\label{fig:Flying-Blimp}Our flying blimp. An autonomous deformable lighter-than-air vehicle (LTAV).}}
}{\small\par}
\end{figure}

\section{Introduction}

In the last decade, the term Unmanned Aerial Vehicles (UAV) has become near-synonymous with so called ``drones'' - small lightweight multirotor craft, kept stable by active control using cheap lightweight MEMS sensors that became ubiquitous as a byproduct of the smartphone revolution \cite{4648819}. Their popularity is mirrored in the field of aerial robotic research. The vast majority of works cover multirotor craft, and to a lesser degree, fixed wing aircraft and helicopters. Thus, flight electronics, control software and simulation tools are readily available for these types of vehicles \cite{EBEID201811,10.5555/1247415.1247446}. This allows researchers to build up a suitable autonomous flying platform for any research task with ease, using commercial off the shelf (COTS) components and open source software.

In contrast, lighter-than-air vehicles (LTAV) such as rigid and nonrigid dirigibles (Fig.~\ref{fig:Flying-Blimp}) have fallen out of fashion, despite their superiority for some applications \cite{10.1007/3-540-45993-6_13}. LTAV are uniquely suited as mobile aerial communication relays, for monitoring and wildlife conservation tasks. They produce little noise, have high energy efficiency and long flight times, display benign collision and crash characteristics, pose low danger and cause little environmental impact. However, their comparably high handling complexity, size, lifting gas requirements and cost create an entry barrier for researchers. Unlike for heavier-than-air ``drones'', there have been no COTS flight controllers that support autonomous dirigible flight. Therefore, guidance and control algorithms have to be implemented for each vehicle - even though various suitable control strategies can be found in the literature \cite{10.1007/3-540-45993-6_13,1272498,4209179,4399531,4543207,770044,770453}. Similar to both rotor craft and fixed wing UAVs, dirigibles come in many types of actuator arrangements: Fixed or vectoring main thrusters, differential thrust, different tail fin arrangements and auxiliary thrusters, single or double hull, etc. Thus, a control algorithm for a specific vehicle might not always be applicable to others.

Development of control algorithms is further complicated by the lack of realistic simulation. Existing robotic simulation environments \cite{EBEID201811} are often ill-suited for the characteristics of LTAV. As LTAV are highly susceptible to wind and turbulence, it is crucial to accurately model both aerodynamic forces and the effects these have on an inherently non-rigid vehicle. To our knowledge, no existing realtime robotic simulation environment addresses this, although analytic studies of these effects have been conducted in the past \cite{LI2011217}.

In this work, we attempt to solve the aforementioned issues in the context of LTAV and provide researchers in the field of aerial robotics with the tools they are accustomed to when working with fixed wing and rotor craft. Our contributions, for which all source code is provided\footnote{Source code: \href{https://github.com/robot-perception-group/airship_simulation}{https://github.com/robot-perception-group/airship\_{}{}simulation}}, are: 
\begin{enumerate}
\item A robotic simulation framework, using ROS/Gazebo \cite{EBEID201811,8548411} for real time hardware in the loop (HITL) and software in the loop (SITL) simulation of easily customizable airship models in a realistic virtual environment. 
\begin{enumerate}
\item Realistic simulation of wind, turbulence \cite{10.5555/887929} and aerodynamic forces on a custom shaped, modular, non-rigid deformable air frame. 
\item Simulated deformation of the air frame in response to aerodynamic, control, thrust and collision forces. 
\item Simulation of buoyancy variation, changes in rigidity and shape, depending on hull pressure and structural parameters. 
\end{enumerate}
\item A baseline guidance and control algorithm for autonomous navigation of airships using hierarchical PI controllers. We employ a generic design, to operate on a large subset of possible airship shape and actuator configurations. 
\begin{enumerate}
\item Implemented based on the Librepilot \cite{EBEID201811,librepilot} open source flight control tool chain, to integrate with existing SITL \& HITL frameworks and available COTS flight control hardware as well as the above mentioned simulation framework. 
\item Full integration with ROS \cite{8548411}. This facilitates easy integration of lighter-than-air vehicles into existing and future robotic research projects for higher level tasks. 
\end{enumerate}
\item Through simulation experiments, we show the effect of changes in rigidity and wind turbulence on the vehicle's controllability, including the introduction of oscillation modes not present in rigid models. 
\item Through real flight experiments, we demonstrate the flight behavior of a $5\mbox{m}$ autonomous blimp UAV, with control coefficients previously determined in our simulation, thus validating our simulation results. 
\end{enumerate}
In Sec.~\ref{sec:State-of-the}, we compare our work to the state of the art. We explain our methodology in Sec.~\ref{sec:Methodology}. Experiments and results are shown in Sec.~\ref{sec:Experiments}, followed by a remark on limitations in Sec.~\ref{sec:Limitations} and our summarizing conclusions in Sec.~\ref{sec:Summary}.

%% file: stateoftheart.tex
\section{State of the art\label{sec:State-of-the}}

The seminal work involving the AURORA airship \cite{10.1007/3-540-45993-6_13,770453,811668} describes the development of a complete system, including novel on board electronics, communication, control algorithms and a flight simulator, both for control evaluation and human operator training. The hierarchical PI control algorithm described there is similar to our implementation, but lacks utilization of dynamic thrust vector control. It also makes many simplifications, including flight at constant airspeed. Although a separate hover control scheme is mentioned, which does use thrust vector control, this is not described in detail. Their simulation assumes rigidity of the air frame. Aerodynamics are modeled based on existing wind tunnel data from a different airship with similar proportions. In contrast, we use modern COTS UAV components and freely available open source software, which greatly enhances reproducibility and the ability to integrate with common robotic components \cite{8548411}. We also model nonrigid deformation effects in our simulation. Our aerodynamic approach is more generic and allows modeling arbitrary dirigibles, even when no detailed aerodynamic measurements are available - based on geometric shape.

The literature review paper \cite{LI2011217} covers modeling of airships, including analysis of deformation and its effect on aerodynamics, spanning 100 years of research, from wind tunnel tests in the early 20th century \cite{jones1924aerodynamical} to computational fluid dynamics (CFD) analysis in the 2000s \cite{lutz2002summary}. We take inspiration from some of the presented methods, especially \cite{jones1983aerodynamic}, which we apply to real time simulation of aerodynamically relevant components. We consider these rigid individually, but allow non-rigid interactions on joints between them. Our simulation does not take aerodynamic cross-interactions between individual components into account. However, we argue that these can be approximated by artificially adjusting lift and drag coefficients to minimize the overall modeling error. This could be done, for example, by comparison to CFD analysis of the rigid shape. However, this is not covered by our current work.

More recent works on airships often involve coupling control with perception and computer vision, such as \cite{1272498,770044,722762,1248908}. All of these works make simplifying assumptions, such as frame rigidity and disregard of wind turbulence.

A novel airship guidance and control scheme was presented in \cite{4543207}, using a backstepping control strategy. Its performance is mathematically proven utilizing a highly simplified analytical plant model under the assumption of rigidity. No real world experiments have been made. This has inspired a number of more recent theoretical works on various airship guidance schemes \cite{liu2020adaptive,cheng2019robust}. However, none of them seem to have been validated in real-world flight. In contrast, we base our model on observed real world flight behavior and validate our results in additional real world flights.

The airship control problem can also be addressed with a model free or learned-model approach \cite{4209179,4399531}. These techniques are well suited to deal with the dynamics of a deforming vehicle under changing conditions from a control perspective, and can be directly applied to the real world. However, a simulation, as presented in our work, is still required to validate and compare the performance of learned algorithms under controlled, reproducible conditions.

%% file: methodology.tex
\section{Methodology\label{sec:Methodology}}

\subsection{Simulation}

Gazebo is a physics simulator often used in robotics, which seamlessly integrates with the Robot Operating System (ROS) \cite{8548411}. Gazebo models rigid physical objects (primitives) based on shape, mass and moments of inertia. Using an XML/URDF description file, these component objects can be described and linked together into hierarchical compound objects. Rigid joints are simulated, using very stiff springs with high damping. These joint parameters can be altered for both linear and rotational freedom. Motion limits, stiffness and dampening properties can be set to simulate non-rigid compound objects. The simulator employs a solver engine to calculate object motions under Newtonian physics, taking all joint forces, collisions and gravity into account. This can be extended through plugins that add additional forces to component objects or joints to model actuators or thrusters. The simulator models time in discrete time steps with configurable temporal resolution and can handle most scenarios in real time with sufficient accuracy. Simulated sensors, which provide simulation data measurements for IMU and positioning sensors as ROS messages are also provided by plugins attached to their respective component objects.

\begin{figure}[t]
\noindent \begin{centering}
{\small{}{}\includegraphics[width=0.49\columnwidth]{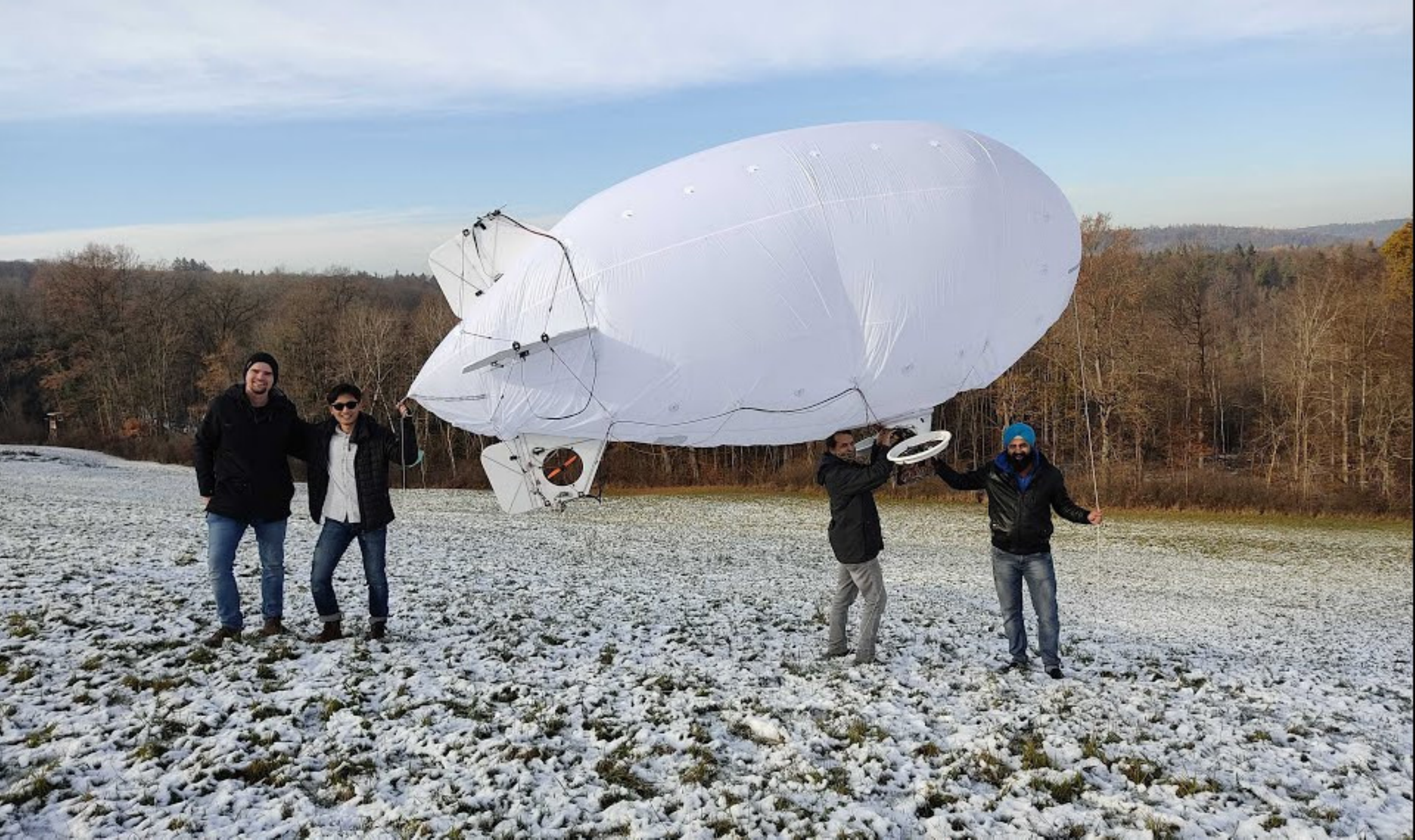}\includegraphics[width=0.49\columnwidth]{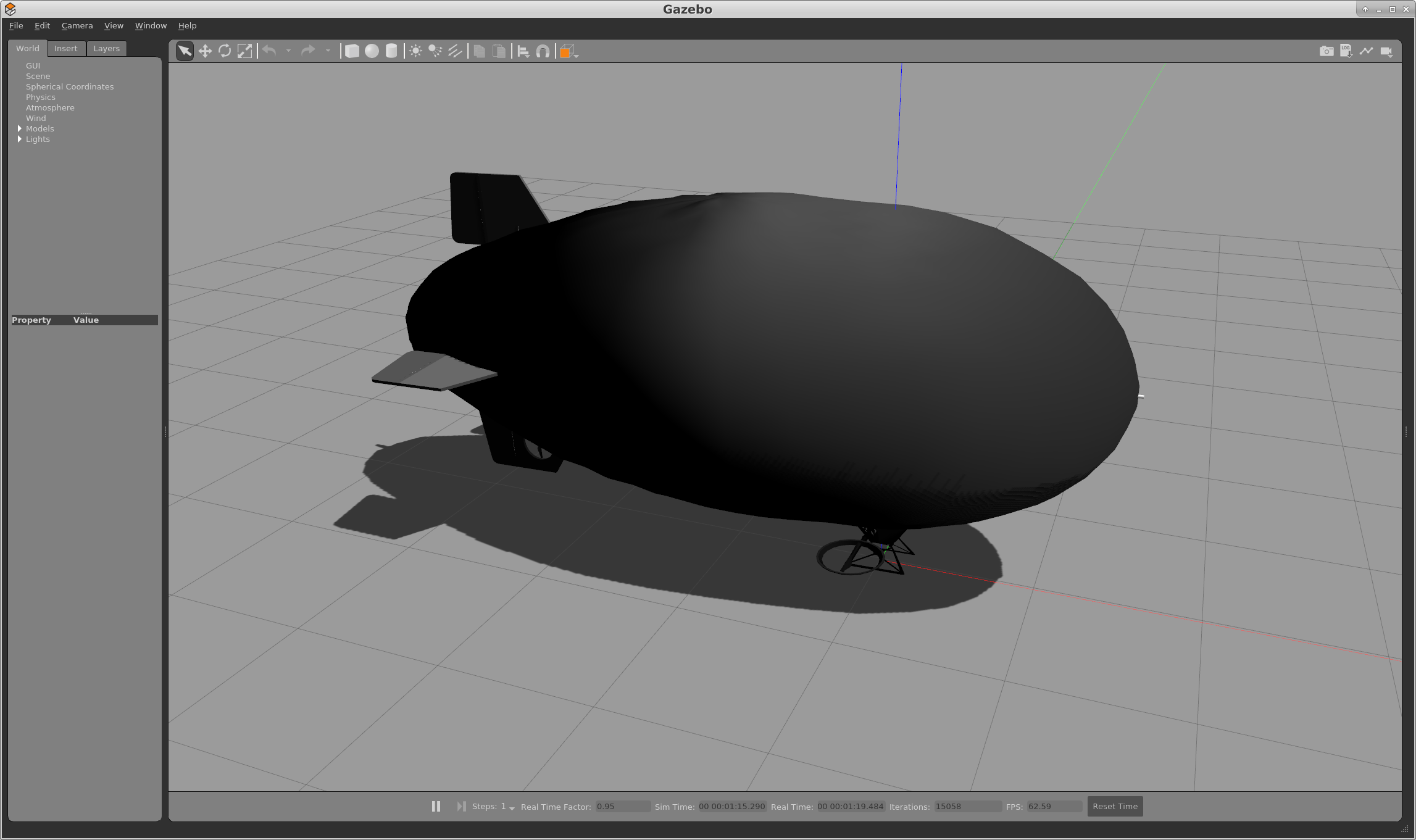}}{\small\par}
\par\end{centering}
\begin{centering}
 
\par\end{centering}
{\small{}{}\caption{{\small{}{}\label{fig:Our-blimp}Our real-world blimp with our team and the corresponding simulated object.}}
}{\small\par}
\end{figure}

We model a blimp (Fig. \ref{fig:Our-blimp}) as a flexible compound object, consisting of rigid primitive component objects for all actuated and non-actuated components. These are fins, rudders, the gondola, thrusters, ballast weights, etc. We model the main hull with at least two separate components, to allow flexibility of the shape as well as higher granularity of forces during rotations. Plugins are added to these primitives to model buoyancy, aerodynamic forces, control actuation and thrust. This setup allows the physical properties of the blimp to be modeled as the sum of its physical components. Mass, dimensions and inertial moments of homogeneous basic shapes can be easily determined by first principles or by measuring the corresponding part on real hardware.{\small{}{}} 

\subsubsection{Buoyancy}

Buoyancy is modeled as a single upward force on sections of the hull based on their volume and coefficients. Lift coefficients can be altered at run-time during the simulation, to experiment with the effects of changing buoyancy.

\subsubsection{Wind}

We model wind as a turbulent flow field, using the Dryden turbulence model (MIL-F-8785C) \cite{10.5555/887929}, the implementation of which is from the FlightGear open source flight simulator \cite{10.5555/1247415.1247446}. It must be noted, that this frequency model simulates both the spacial and temporal variability of turbulence in a single, moving point only. As such, we can only calculate a single, time variable global flow vector, which we assume constant in space; the same assumption that was made by FlightGear. For small airships we deem this is an acceptable limitation. This flow vector is published as a ROS message, so it is available to other plugins at any point in time.

\subsubsection{Lift and drag}

For all aerodynamically relevant primitive objects, including hull sections, we calculate the local lift and drag forces based on the orientation and motion of the object through the flow field, defined by the current wind flow vector. These are then taken into account by the physics solver to model the effect on the whole vehicle as well as its deformations. We model two types of basic lifting primitives, ``quasi-planar'' and ``quasi-cylindrical''. Let $\overline{f}$ be the flow vector relative to the primitive, consisting of orthogonal components, $f_{x}$ from the front, $f_{y}$ from the left and $f_{z}$ from above. For quasi-planar types $f_{y}$ is ignored. Quasi-cylindrical objects are rotated around their $x$ (forward) axis. This yield a rotated flow $_{r}\overline{f}$ such tha{\small{}{}t 
\begin{equation}
_{r}\overline{f}=\left[f_{x},0,\left|f_{y}\right|+\left|f_{z}\right|\right]
\end{equation}
}From that point on, they are treated identical to quasi-planar types. We calculate the angle of attack $\alpha$, the normalized drag vector $\hat{f}_{d}$, normalized lift vector $\hat{f}_{l}$ and dynamic pressure $q_{d}$ a{\small{}{}s 
\begin{equation}
\alpha=\arctan\left(\frac{f_{z}}{f_{x}}\right)\mbox{ ,}
\end{equation}
\begin{equation}
\hat{f}_{d}=\frac{\overline{f}_{d}}{\left|\overline{f}_{d}\right|}\mbox{ for }\overline{f}_{d}=\left[f_{x},0,f_{z}\right]\mbox{ ,}
\end{equation}
\begin{equation}
\hat{f}_{l}=\hat{f}_{d}\times\left[0,\frac{\alpha}{\left|\alpha\right|},0\right]\mbox{ ,}
\end{equation}
\begin{equation}
q_{d}=k\left(f_{x}^{2}+f_{z}^{2}\right)\mbox{ .}
\end{equation}
}Let $A$ be the relevant area of the object, given coefficients $c_{l_{0}}$, $c_{d_{0}}$, $c_{d_{1}}$ and $\alpha_{stall}$. We calculate forces for lift $\overline{F}_{l}$ and drag $\overline{F}_{d}$ as

{\small{}{} 
\begin{equation}
\overline{F}_{l}=q_{d}A\hat{f}_{l}c_{l}\mbox{ , }c_{l}=c_{l_{0}}\begin{cases}
\frac{\left|\alpha\right|}{\alpha_{stall}} & \mbox{if }\left|\alpha\right|\leq\alpha_{stall}\\
\frac{\pi-2\left|\alpha\right|}{\pi-2\alpha_{stall}} & \mbox{if }\left|\alpha\right|>\alpha_{stall}
\end{cases}\mbox{ ,}
\end{equation}
}{\small\par}

\begin{figure}[t]
\noindent \begin{centering}
{\small{}{}\includegraphics[width=0.5\columnwidth]{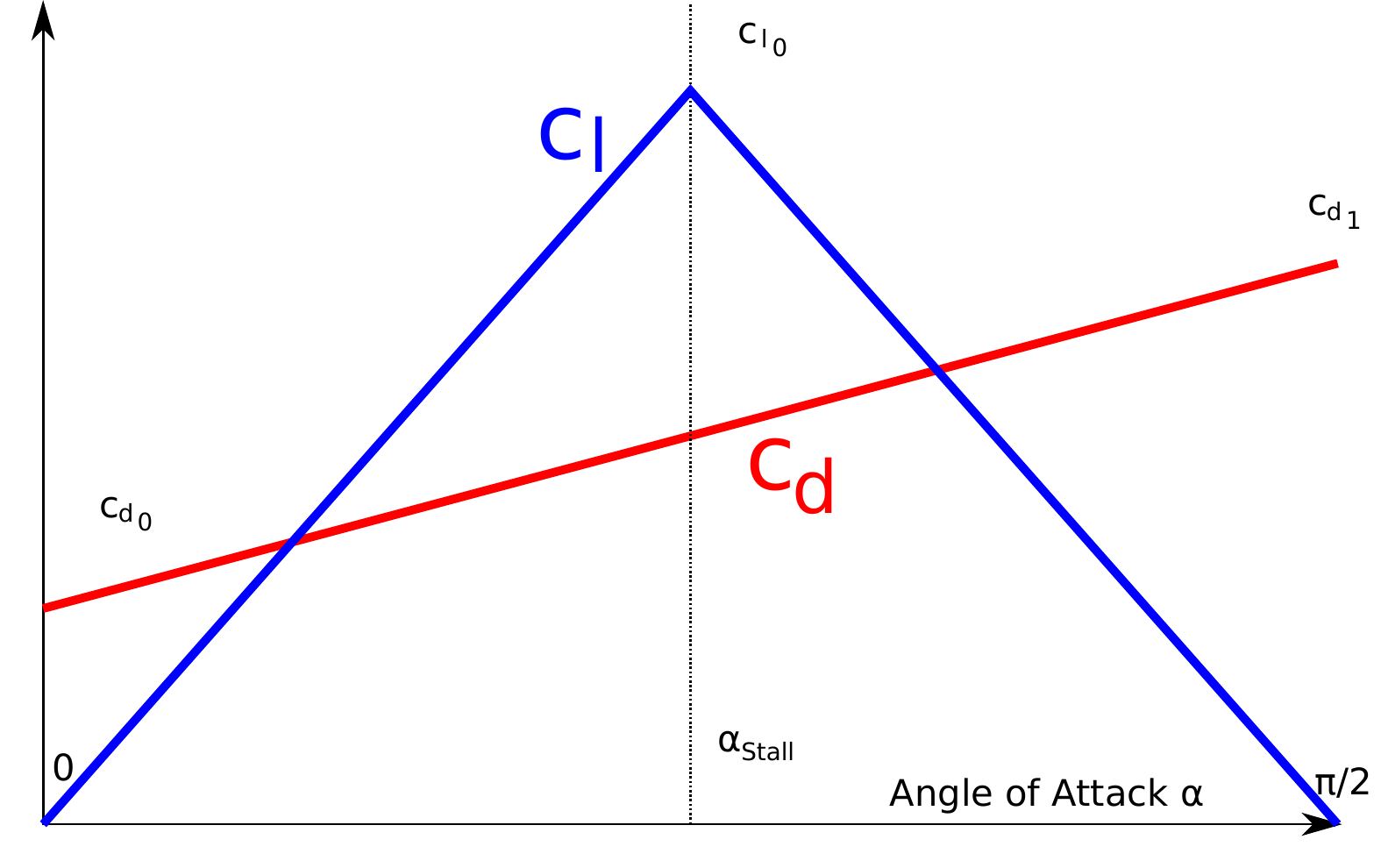}\includegraphics[width=0.5\columnwidth]{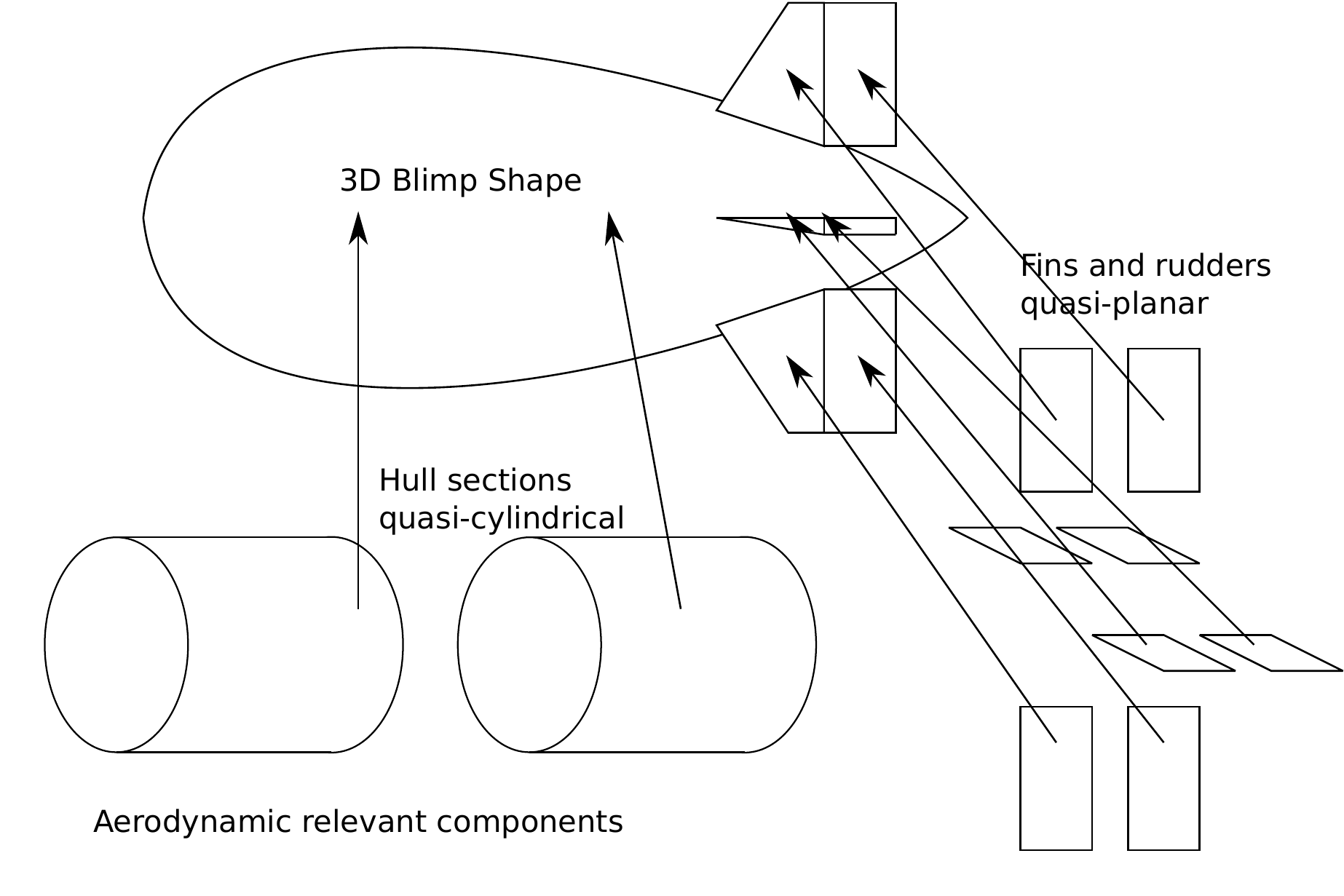}}{\small\par}
\par\end{centering}
\begin{centering}
 
\par\end{centering}
{\small{}{}\caption{{\small{}{}\label{fig:LiftAndDrag}Functions for lift and drag coefficients for a simulated object component. These are calculated for each aerodynamically relevant primitive component of the blimp.}}
}{\small\par}

\end{figure}

{\small{}{} 
\begin{equation}
\overline{F}_{d}=q_{d}A\hat{f}_{d}c_{d}\mbox{ , }c_{d}=c_{d_{0}}\left(1-\frac{2\left|\alpha\right|}{\pi}\right)+c_{d_{1}}\frac{2\left|\alpha\right|}{\pi}\mbox{ .}
\end{equation}
}Both functions are highly simplified linear approximations (Fig. \ref{fig:LiftAndDrag}) of the true lift and drag curves. For each primitive object, only 4 aerodynamic coefficients are needed. 

The approximate drag coefficients for basic primitive shapes, both for frontal and sideways flow are commonly known, as are lift coefficients of thin rectangular plates or cylinders. Care has to be taken to adjust the coefficients when nearby primitives interact with each other on a compound object. For example, only the first and last section of the hull would have a high $c_{d_{0}}$ due to pressure drag, while any intermediate sections would only experience friction drag which typically is an order of magnitude lower. Due to the linear contribution, assuming identical $A$ for $n$ hull sections and frontal flow drag coefficient $_{H}c_{d}$ for the basic shape, we observe {\small{}{} 
\begin{equation}
\underset{k=1}{\overset{n}{\sum}}{}_{k}c_{d_{0}}={}_{H}c_{d}\mbox{ ,}
\end{equation}
} and set the coefficients for the hull sections accordingly, weighted by their expected drag contribution.

As another example, fins attached to the hull might encounter increased lift due to increased dynamic pressure of the airflow around the hull, which can be modeled by artificially inflated lift coefficients for these surfaces.

If an aerodynamic model exists for a specific vehicle, it should be possible to solve for a ``best fit'' of the primitive coefficients. However, our approach has the advantage that a roughly approximated aerodynamic model can be created based on geometric shape only, which we argue is a valuable property for evaluating new vehicle designs.

\subsubsection{Non-rigidity}

\begin{figure}
\noindent \begin{centering}
{\small{}{}\includegraphics[width=0.26\columnwidth]{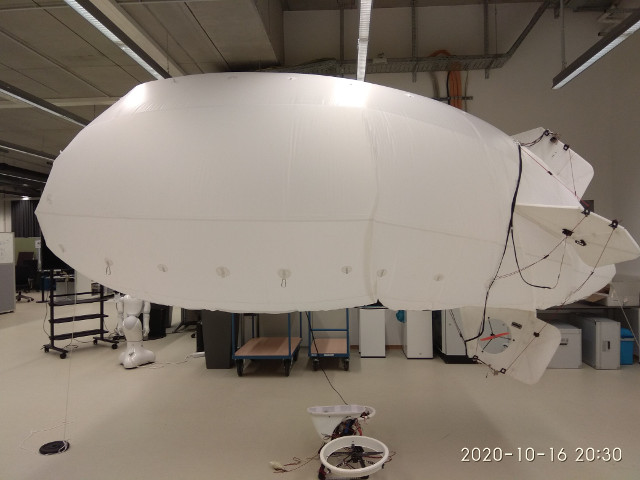}\includegraphics[width=0.146\columnwidth]{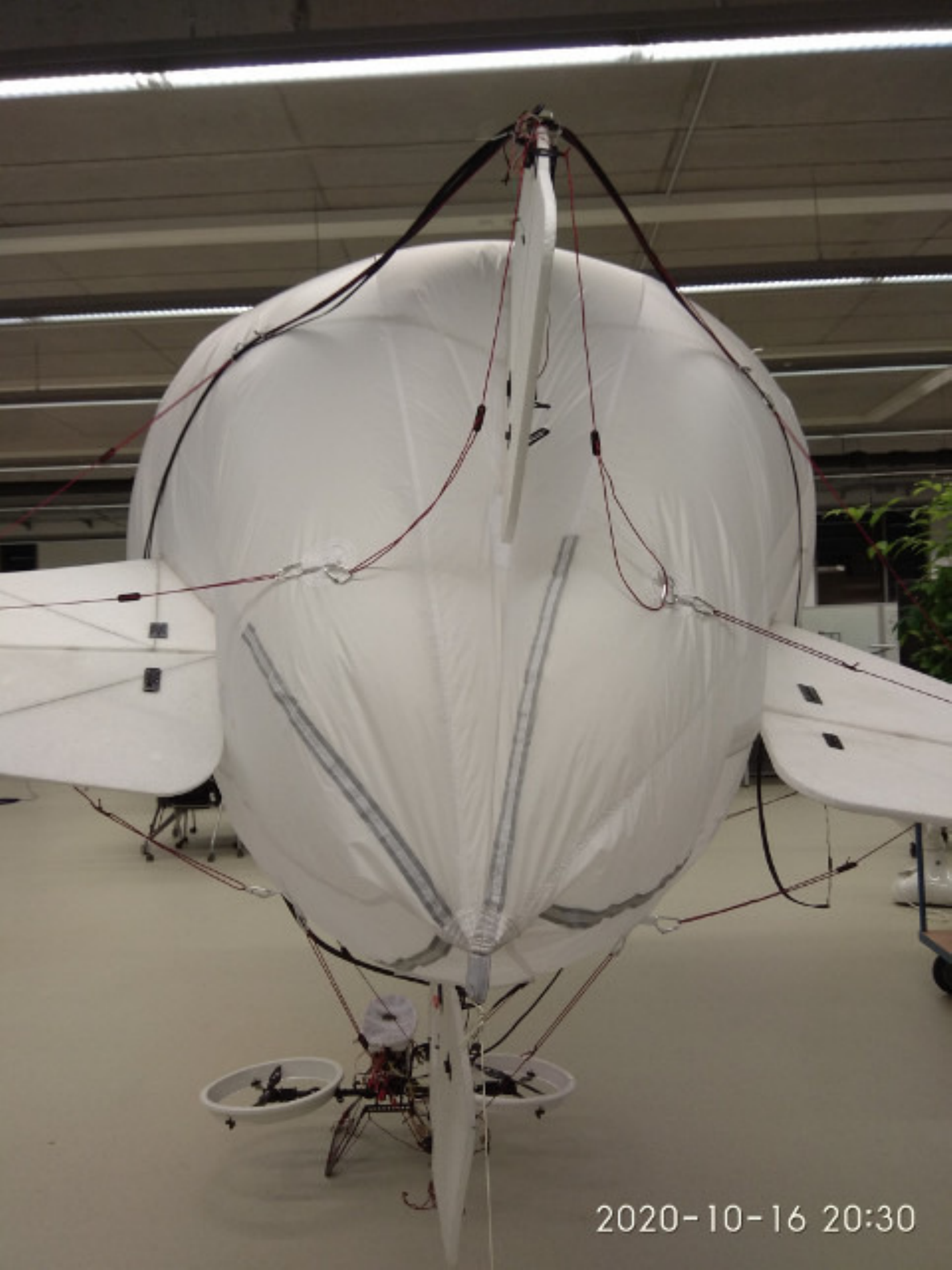}\includegraphics[width=0.292\columnwidth]{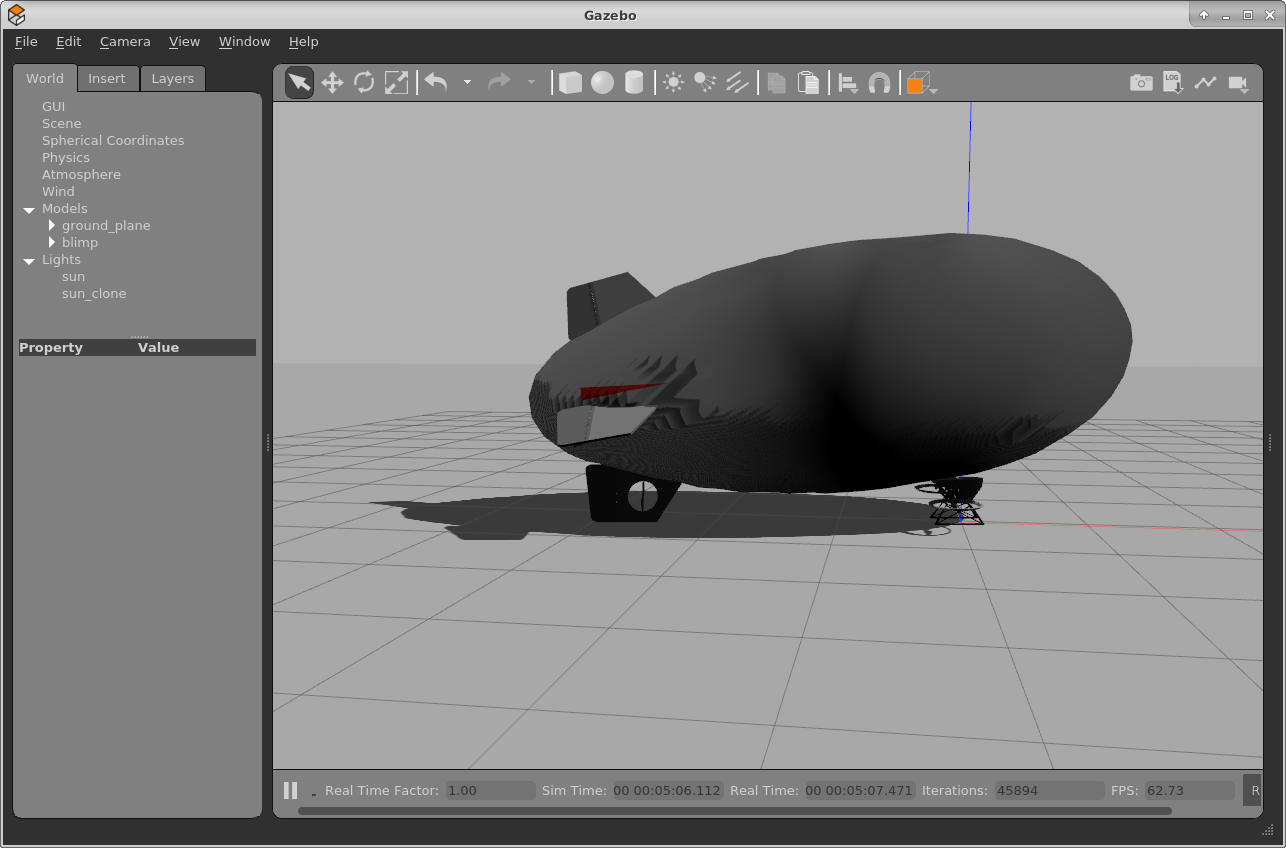}\includegraphics[width=0.292\columnwidth]{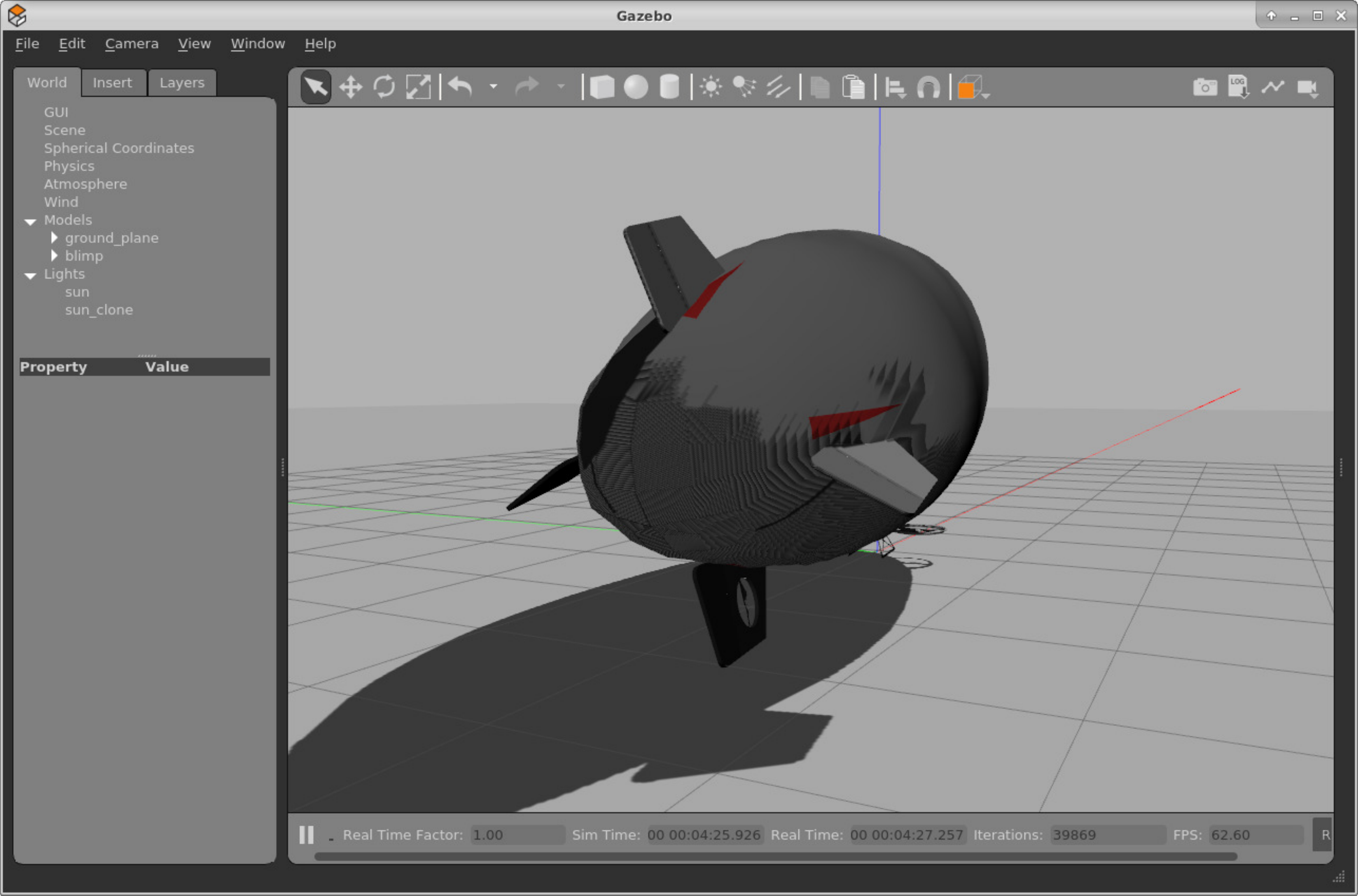}}{\small\par}
\par\end{centering}
\begin{centering}
 
\par\end{centering}
{\small{}{}\caption{{\small{}{}\label{fig:Deflation-pic}Deflation of the blimp affects its shape, both in reality and simulation. The hull is visually rendered rigid, but aerodynamically it is simulated in sections. Under their weight, the fins sink into the hull and drag the tail down until the bottom fin carries part of the blimp's weight. These images show severe deflation for better visibility. Flight in this state would not be possible.}}
}{\small\par}

\end{figure}

Gazebo joint properties in compound objects can be manipulated through a ROS service while a simulation is running. We wrote a script that can simultaneously adjust both the rotational and linear spring stiffness, as well as the range of free rotation for all joint connections between the blimp hull and other components such as fins and gondola around all axis of rotation. The same program also adjusts the buoyancy of the blimp. This way, the effects of a drop or increase in lifting gas pressure and volume can be evaluated at any time, both regarding flight behavior and controllability (Fig.~\ref{fig:Deflation-pic}). 

\subsection{Control}

We employ a cascaded PI controller design similar to the controller presented in \cite{10.1007/3-540-45993-6_13,770453}, with extensions. Librepilot \cite{librepilot,EBEID201811} offers a vehicle-agnostic control hierarchy based on virtual control axes, which are mapped to physical actuators through a configurable matrix. This maps a ``Pitch'' virtual axis on the elevator servos, the ``Yaw'' virtual axis on both rudder servos and a yaw thruster, as well as a ``Thrust'' virtual axis on both main thrusters. We decided not to employ main thruster differential thrust for control, due to the low torsional moment and high strain inflicted between gondola and hull. A simple PI control loop controls the yaw rate, while a hierarchical controller with an outer loop proportional term controls the pitch angle based on an PI inner loop controlling pitching rate. Implementations for these already exist and are utilized both for fixed wing and multirotor control.

In extension, we implemented a novel path follower, which determines the setpoints for the lower level control: Pitch $P$, Yaw rate $\dot{Y}$, Thrust $T$ and Thrust vector angle $0\leq\gamma\leq\frac{\pi}{2}$. Its inputs are a setpoint velocity vector in 3d space $\overline{v}_{S}$, the airspeed $v_{I}$, the velocity in 3d space $\overline{v}=\left[v_{x},v_{y},v_{z}\right]$ and the current attitude of the vehicle in Euler angles $q=\left[q_{r},q_{p},q_{y}\right]$. All the state estimates come from the flight controller's state estimation using an extended Kalman filter (EKF). If no airspeed sensor is available, $v_{I}$ is assumed to be the forward component of $\overline{v}$. We compensated the wind to calculate motion relative to the flow field. Let $q\left(v_{I}\right)$ be a function that rotates $v_{I}$ into the world-frame. We then calculate the estimated wind flow vector $\overline{f}_{l}$ and the corrected velocity in air $\tilde{\overline{v}}$ as{\small{}{} 
\begin{equation}
\overline{f}_{l}=\overline{v}-q\left(v_{I}\right)\mbox{ ,}
\end{equation}
\begin{equation}
\tilde{\overline{v}}=\left[\tilde{v}_{x},\tilde{v}_{y},\tilde{v}_{z}\right]=\overline{v}-\overline{f}_{l}\mbox{ .}
\end{equation}
}For each vehicle a minimum controllable airspeed $v_{min}$ and a maximum achievable airspeed $v_{max}$ are defined. We solve for a wind-corrected setpoint $\tilde{\overline{v}}_{S}$ such that{\small{}{} 
\begin{equation}
\tilde{\overline{v}}_{S}=\left[\begin{array}{c}
\tilde{v}_{sx}\\
\tilde{v}_{sy}\\
\tilde{v}_{sz}
\end{array}\right]=b\overline{v}_{s}-\overline{f}_{l}\mbox{ , }\begin{cases}
b>0\mbox{, }\left|b\overline{v}_{s}-\overline{f}_{l}\right|=v_{min} & \mbox{if }\left|\overline{v}_{S}-\overline{f}_{l}\right|<v_{min}\\
b=1 & \mbox{if }v_{min}\leq\left|\overline{v}_{S}-\overline{f}_{l}\right|\leq v_{max}\\
b>0\mbox{, }\left|b\overline{v}_{s}-\overline{f}_{l}\right|=v_{max} & \mbox{if }\left|\overline{v}_{S}-\overline{f}_{l}\right|>v_{max}
\end{cases}\mbox{ .}
\end{equation}
}If the wind is stronger than the vehicle's maximum speed, there might be no solution with $b>0$. In this case, the solution with the highest $b$ is chosen, which minimizes the positional error accumulated under these conditions.

Separate controllers are employed for direction $C_{d}$, airspeed $C_{v}$, climb rate $C_{\dot{h}}$and thrust vector $C_{\gamma}$. Let $\varangle\left(\overline{a},\overline{b}\right)$ be the signed angular difference between $\overline{a}$ and $\overline{b}$ while $\int_{\left\langle \pm l_{i}\right\rangle }$ shall denote an integral accumulator limited by $\pm l_{i}$. We then calculate{\small{}{} 
\begin{equation}
\dot{Y}=C_{d}\left(\tilde{\overline{v}}_{S},\tilde{\overline{v}}\right)={}_{\dot{Y}}k_{p}\varangle\left(\left[\tilde{v}_{sx},\tilde{v}_{sy}\right],\left[\tilde{v}_{x},\tilde{v}_{y}\right]\right)\mbox{ ,}
\end{equation}
\begin{equation}
P=C_{\dot{h}}\left(\tilde{\overline{v}}_{S},\tilde{\overline{v}}\right)={}_{P}k_{p}\left(\tilde{v}_{sz}-\tilde{v}_{z}\right)+\int_{\left\langle \pm_{P}l_{i}\right\rangle }\left(_{P}k_{i}\left(\tilde{v}_{sz}-\tilde{v}_{z}\right)dt\right)\mbox{ ,}
\end{equation}
\begin{equation}
T=C_{v}\left(\tilde{\overline{v}}_{S},\tilde{\overline{v}}\right)={}_{T}k_{p}\left(\left|\tilde{\overline{v}}_{S}\right|-\left|\tilde{\overline{v}}\right|\right)+\int_{\left\langle \pm_{T}l_{i}\right\rangle }\left(_{T}k_{i}\left(\left|\tilde{\overline{v}}_{S}\right|-\left|\tilde{\overline{v}}\right|\right)dt\right)+{}_{P\times T}k_{p}P\mbox{ ,}
\end{equation}
\begin{equation}
\gamma=C_{\gamma}\left(\tilde{\overline{v}}_{S},\tilde{\overline{v}}\right)={}_{\gamma}k_{p}P\mbox{ .}
\end{equation}
}{\small\par}

The main addition to \cite{10.1007/3-540-45993-6_13,770453} is the thrust vector term $C_{\gamma}$ which, in combination with the proportional climb speed cross term $_{P\times T}k_{p}P$ in $C_{v}$, employs the main thrusters for altitude control. For a vehicle without thrust vector control, the corresponding coefficients $_{\gamma}k_{p}$, $_{P\times T}k_{p}$ would remain $0$.

\subsection{Middleware}

The flight control algorithm is implemented in C/C++ and executed either on a physical embedded flight controller (Openpilot Revolution) \cite{EBEID201811} on the LTAV, both in flight or for HITL simulation, or compiled as a computer program for SITL simulation. The Librepilot Ground Control Software (GCS) connects to either via telemetry/networking and offers a HITL/SITL interface, in which the flight controller's sensor measurements are overridden with simulated data and actuator commands are sent to the simulator. We extended this interface, in order to connect with ROS, and as such our Gazebo simulation. We also wrote additional software that interfaces with the flight controller directly (via USB or Networking) and allows HITL/SITL without the GCS. Its main purpose however, is to send sensor and state estimate data to ROS components running on board and in flight. In turn ROS can send the flight controller guidance setpoints. This facilitates high level autonomous control, such as vision or perception based algorithms, through ROS. Of course, these components are useful beyond their application for airships.

\subsection{Robotic Hardware}

We equipped a commercial $5\mbox{m}$ blimp (Fig.~\ref{fig:Our-blimp}) (CloudMedia Sopot, Sopot, Poland), designed for advertising and aerial photography, with an Openpilot Revolution flight controller (including an IMU, 3 axis magnetometer, barometer and processor), a GPS receiver, a digital airspeed sensor, and an onboard computer for data logging and wireless networking. All sensors and the flight controller have been installed on the top vertical tail fin, to avoid electromagnetic interference. The main computer, a NVIDIA Jetson TX1, is installed in the gondola and connected with a lightweight USB cable. Telemetry is relayed to the ground via 5 GHz WiFi using the main computer's inbuilt transceiver. The blimp can be flown manually using a Graupner flight control transmitter (TX), a matching receiver is connected to the flight controller. Autonomous flight modes are engaged using switches on the TX. The blimp has 8 independently controllable actuators: i) Bottom fin lateral thruster (propeller with forward and reverse thrust), ii) Top rudder (servo), iii) Bottom rudder (servo), iv) Left elevator (servo), v) Right elevator (servo), vi) Thrust vector control axis (servo), vii) Left main thruster (forward and reverse thrust), viii) Right main thruster (forward and reverse thrust). This vehicle has a mass of $10\mbox{kg}$ including a payload or ballast of approximately $1\mbox{kg}$.

%% file: experiments.tex
\section{Experiments and Results\label{sec:Experiments}}

We conducted several experiments, both in simulation as well as real-world validation.

\subsection{Simulation experiments}

\subsubsection{Simulation experiment 1 - manual simulated flight}

\noindent - Based on data from physical measurements and observations of a tethered real-world test flight\footnote{Video, test flight: \href{https://youtu.be/ZTCQNS4tmqo}{https://youtu.be/ZTCQNS4tmqo}}, the simulation model was configured and tested in manual simulated flight. Unknown coefficients were adjusted until simulation and reality were in qualitative agreement\footnote{Video, simulation experiment 1: \href{https://youtu.be/LNybtK3HNp8}{https://youtu.be/LNybtK3HNp8}}.

\subsubsection{Simulation experiment 2 - loitering}

\begin{figure}
\noindent \begin{centering}
{\small{}{}\includegraphics[width=0.49\columnwidth]{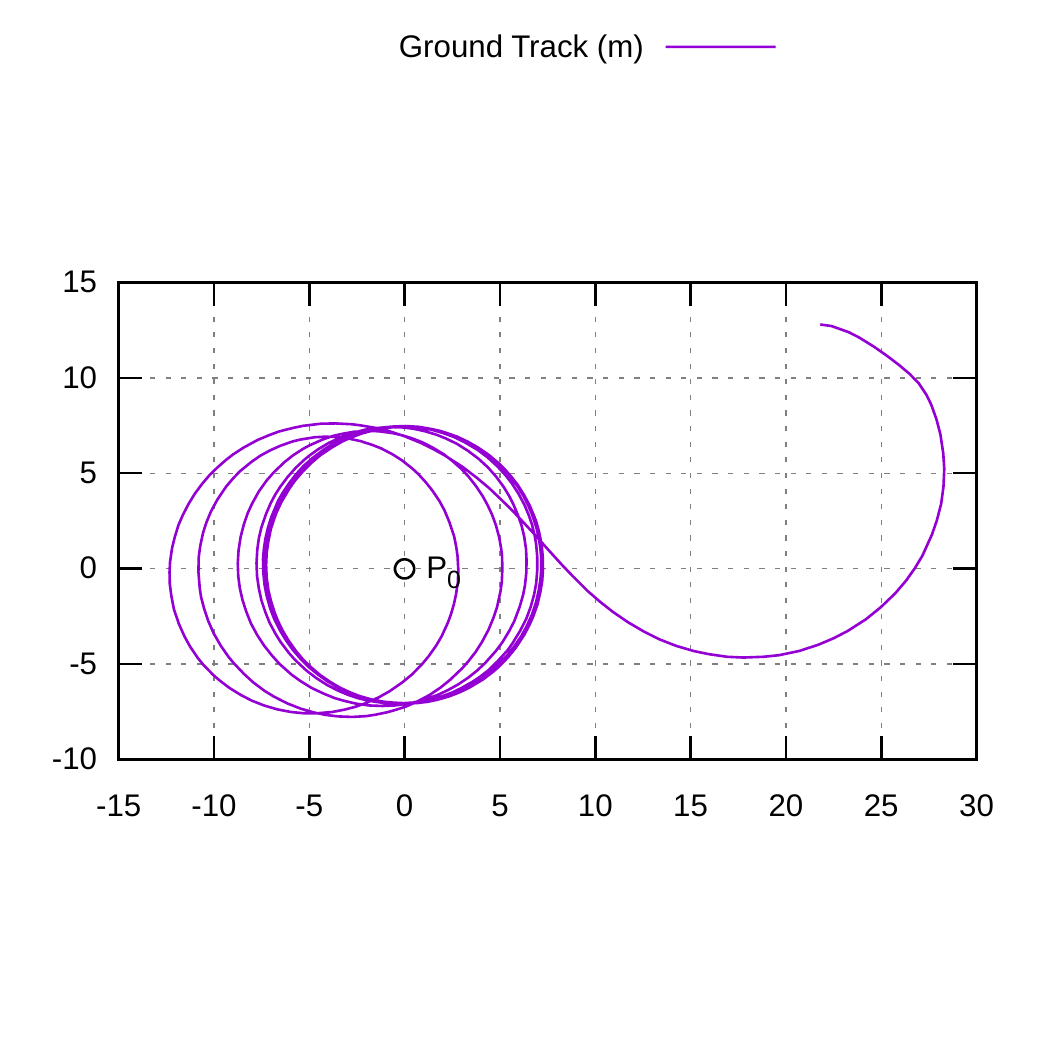}\includegraphics[width=0.49\columnwidth]{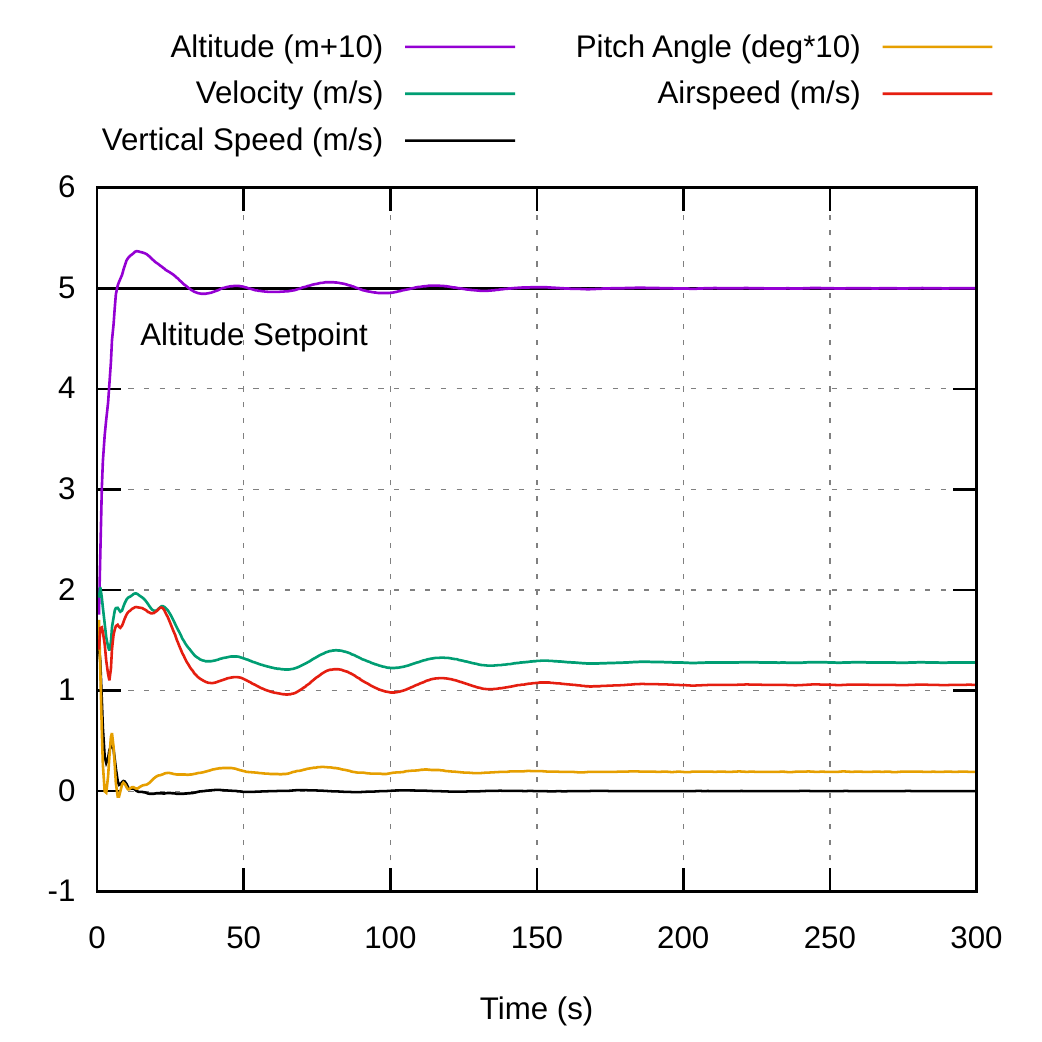}}{\small\par}
\par\end{centering}
\begin{centering}
 
\par\end{centering}
\caption{{\small{}{}\label{fig:Sim exp 2}}\textbf{\small{}{}Simulation experiment 2}{\small{}{} - The simulated blimp loiters around a given position. Without wind, the steady state becomes a circle at airspeed $v_{min}=1\mbox{m/s}$ and at the minimum possible turn radius. The turn rate throughout all experiments was limited at $\pm10\nicefrac{\mbox{\textdegree}}{\mbox{s}}$.}}
\end{figure}

\noindent - The control algorithm was implemented and control parameters were tuned in simulation, starting with the lowest level control loop. Given manual setpoint inputs, P and I rate control coefficients were determined for pitch and yaw rate, followed by P coefficient for the pitch control loop. When these were satisfactory, the $C_{d}$ control loop was tuned, followed by $C_{\dot{h}}$, $C_{v}$ and $C_{\gamma}$. We then followed an iterative tuning procedure, while the simulated blimp was constantly loitering around a position $P_{0}$. Given current position estimate $P$, the velocity setpoint is calculated as {\small{}{} 
\begin{equation}
\overline{v}_{S}=k\left(P_{0}-P\right)\mbox{ .}
\end{equation}
}With the simulated blimp loitering, the performance in each controlled domain was monitored and control coefficients for whichever parameter (speed, altitude, course) was deemed worst were adjusted until satisfactory results were achieved (Fig.~\ref{fig:Sim exp 2}). 

\subsubsection{Simulation experiment 3 - trajectory following}

\noindent - We employ the path planner in Librepilot to follow a waypoint sequence with 4 waypoints. We denote $\hat{p}=\frac{\overline{p}}{\left|\overline{p}\right|}$ as the notation for a normalized vector. For any pair of consecutive waypoints $P_{0}$, $P_{1}$ and desired velocity $v_{0}$, we determine $\overline{v}_{S}$ via

\noindent {\small{}{} 
\begin{equation}
\overline{p}_{a}=P-P_{0}\mbox{ ,}
\end{equation}
\begin{equation}
\overline{p}_{b}=P_{1}-P_{0}\mbox{ ,}
\end{equation}
}{\small\par}

\begin{figure}
\noindent \begin{centering}
{\small{}{}\includegraphics[width=0.49\columnwidth]{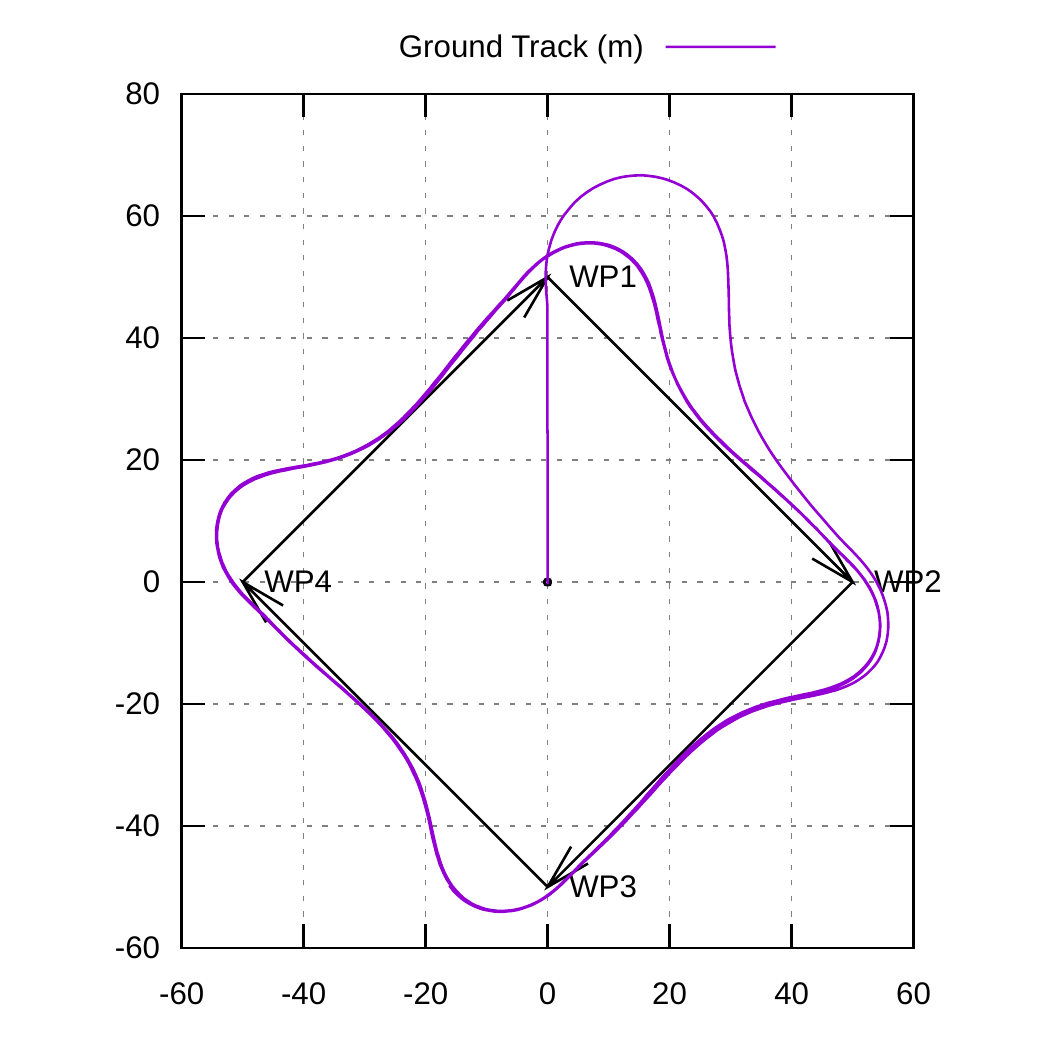}\includegraphics[width=0.49\columnwidth]{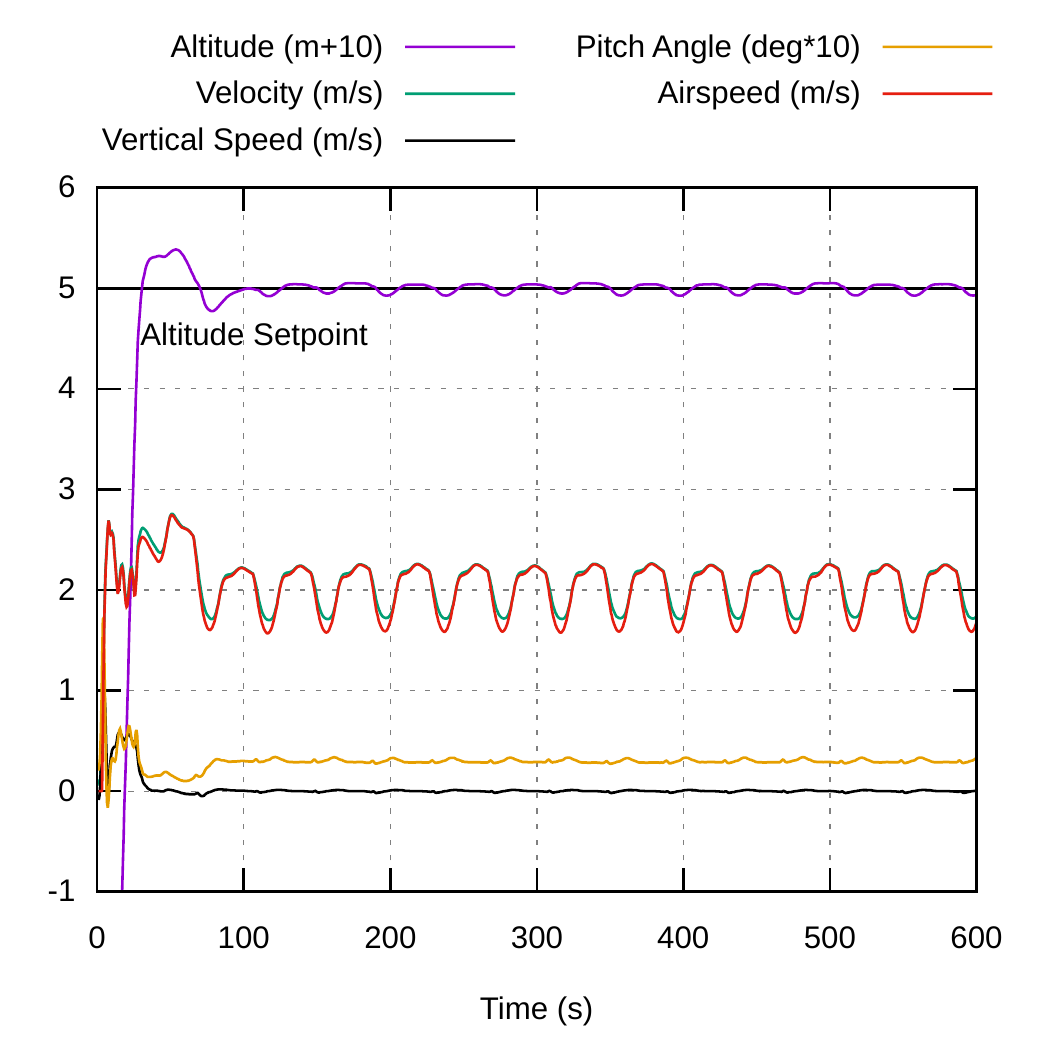}}{\small\par}
\par\end{centering}
\begin{centering}
 
\par\end{centering}
\caption{{\small{}{}\label{fig:Sim exp 3}}\textbf{\small{}{}Simulation experiment 3}{\small{}{} - The simulated blimp follows a waypoint sequence at a commanded speed of $2\mbox{m/s}$.}}
\end{figure}

{\small{}{} 
\begin{equation}
\overline{v}_{S}=v_{0}\hat{p}_{b}+k\left(\overline{p}_{b}\left(\overline{p}_{a}\cdot\hat{p}_{b}\right)-\overline{p}_{a}\right)\mbox{ .}\label{eq:19}
\end{equation}
}In equation (\ref{eq:19}) the term $v_{0}\hat{p}_{b}$ moves the vehicle along the desired path vector, while the second term corrects any orthogonal deviance from the path, as $\overline{p}_{b}\left(\overline{p}_{a}\cdot\hat{p}_{b}\right)$ projects $\overline{p}_{a}$ on $\overline{p}_{b}$. This guidance scheme allows observations of the control behavior in steady state, while the vehicle follow a reproducible trajectory repeatedly (Fig. \ref{fig:Sim exp 3}). 

\subsubsection{Simulation experiment 4 - wind and turbulence}

\begin{figure}
\noindent \begin{centering}
{\small{}{}Loitering:}\\
 {\small{}{}\includegraphics[width=0.49\columnwidth]{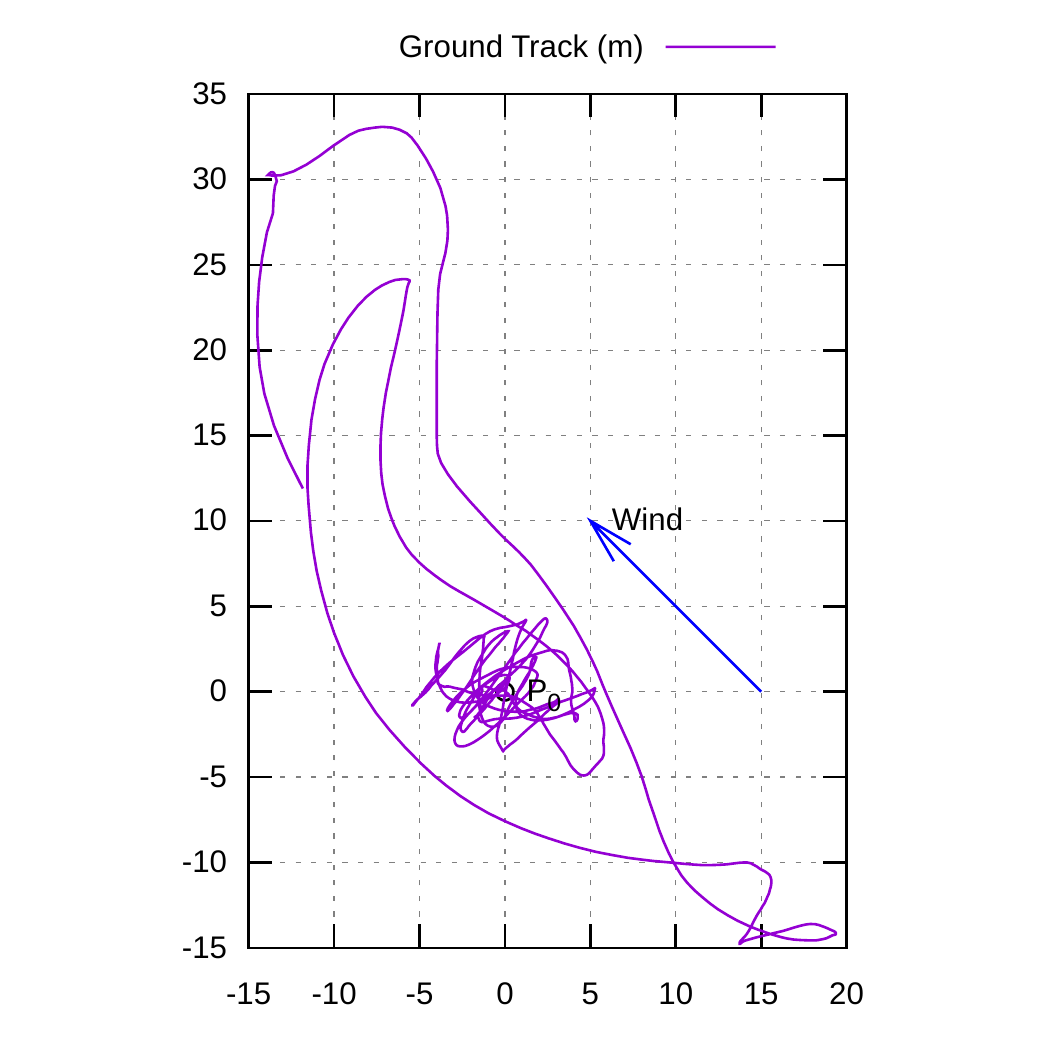}\includegraphics[width=0.49\columnwidth]{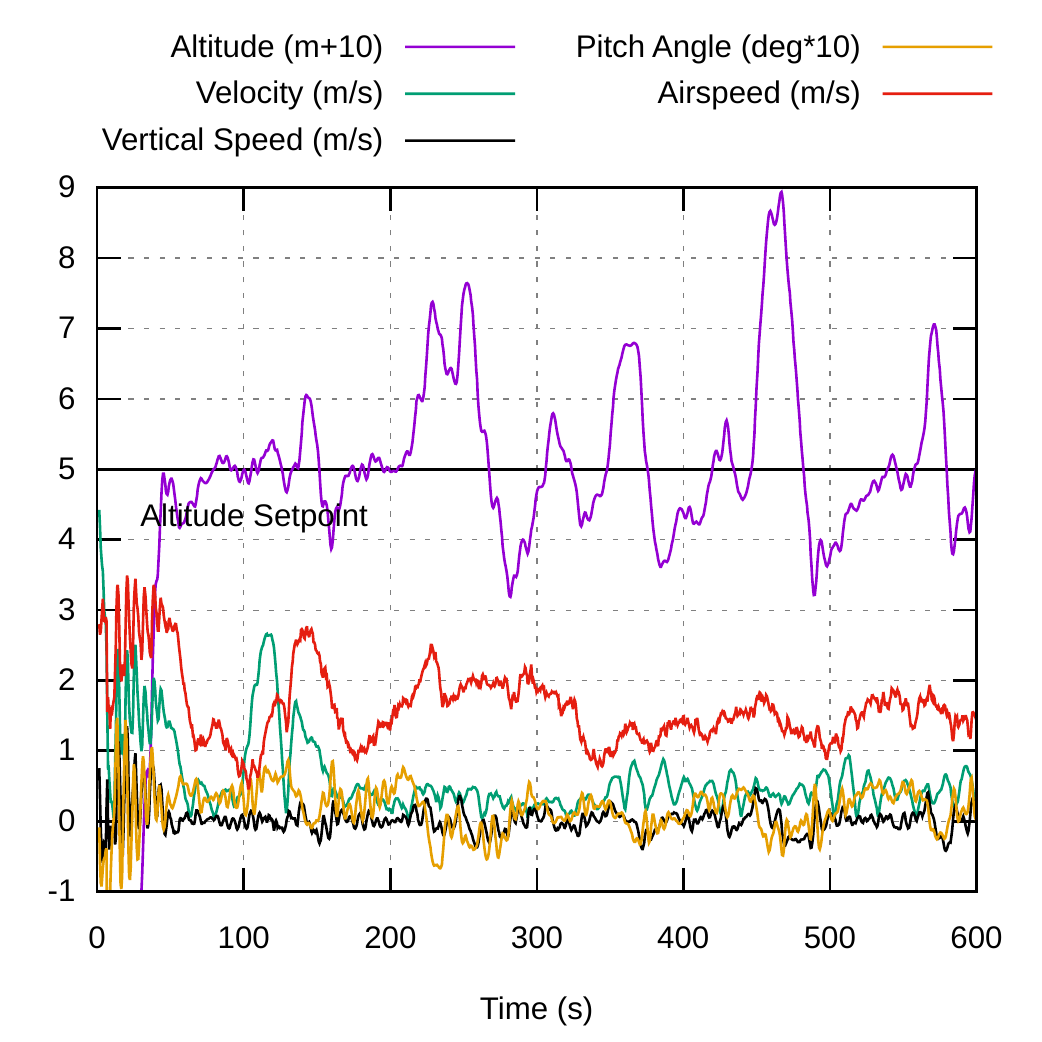}}\\
 {\small{}{}Trajectory Following:}\\
 {\small{}{}\includegraphics[width=0.49\columnwidth]{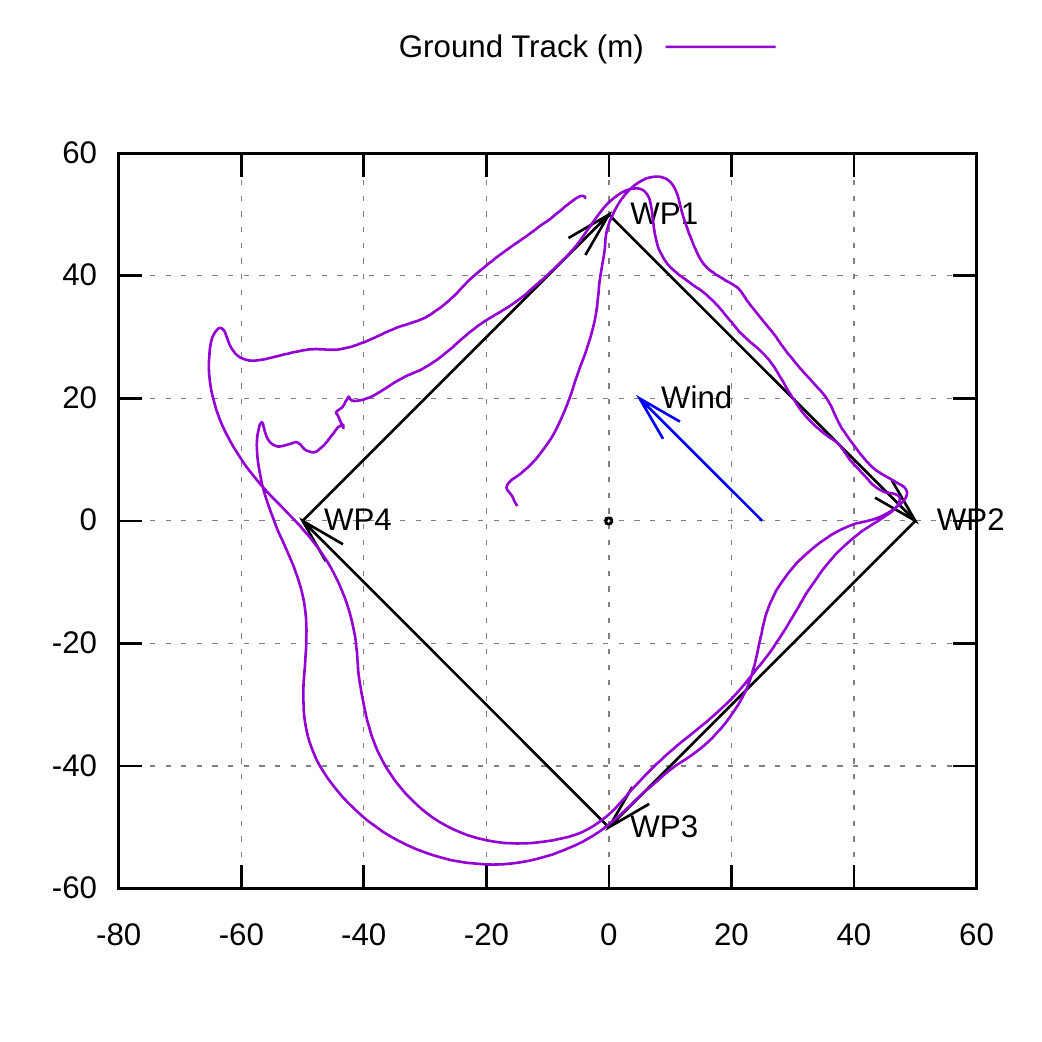}\includegraphics[width=0.49\columnwidth]{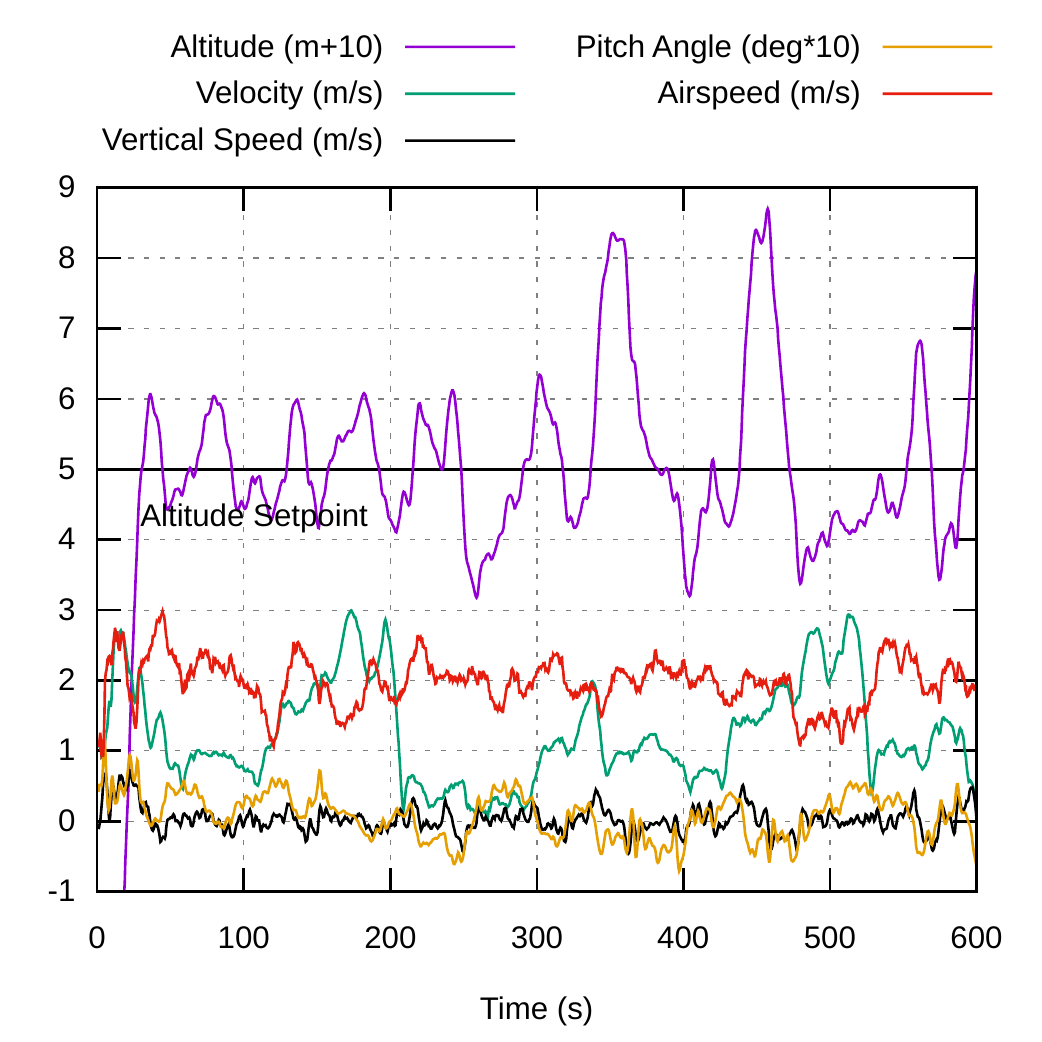}}{\small\par}
\par\end{centering}
\begin{centering}
 
\par\end{centering}
\caption{{\small{}{}\label{fig:Sim exp 4}}\textbf{\small{}{}Simulation experiment 4}{\small{}{} - Navigating in wind poses a challenge for the simulated blimp, as the simple action of turning around after an overshoot leads to significant drift downwind. However, with the wind speed between $v_{min}=1\mbox{m/s}$ and $v_{max}=2\mbox{m/s}$ the blimp manages to hover in close proximity of the reference point. The ``random walk'' is an effect of turbulent fluctuations in wind speed and direction. When navigating a waypoint sequence, the ground speed is severely increased on the downwind leg, even at an airspeed of $v_{min}$, resulting in large turn radii and overshoot. On the upwind leg the ground speed is greatly reduced, coming to a near halt during strong gusts. The gusty wind poses challenges to the blimp's ability to maintain altitude, vertical gust components cause severe altitude excursions of several meters which the blimp then compensates to the best of its abilities. These are reproducible regardless of the guidance mode, leading to similar excursions at the same time in each experiment, whenever the same wind gusts are encountered. Simulated wind is coming from south-east ($135\mbox{\textdegree}$) with an average velocity of $1.5\mbox{m/s}$, simulated turbulence for the Dryden model was set to magnitude $3$.}}
\end{figure}

\noindent - We added wind and turbulence to situations from both simulation experiment 2 and experiment 3 and observe the effects. Wind velocities between $v_{min}$ and $v_{max}$ allowed the simulated blimp to hover on the spot with good accuracy. We noted altitude excursions of several meters due to vertical wind gust components. During the trajectory following, wind from the rear poses a challenge if $v_{min}$ is set too low, as the the blimp is almost stationary in relation to the air-stream. Consequently, the control surfaces have very limited authority. This led to a noticeable overshoot of waypoints due to the time it took for the blimp to reorient. This can be alleviated by commanding a higher $v_{min}$ (Fig. \ref{fig:Sim exp 4}). 

\subsubsection{Simulation experiment 5 - deflation}

\begin{figure}
\noindent \begin{centering}
{\small{}{}Loitering:}\\
 {\small{}{}\includegraphics[width=0.49\columnwidth]{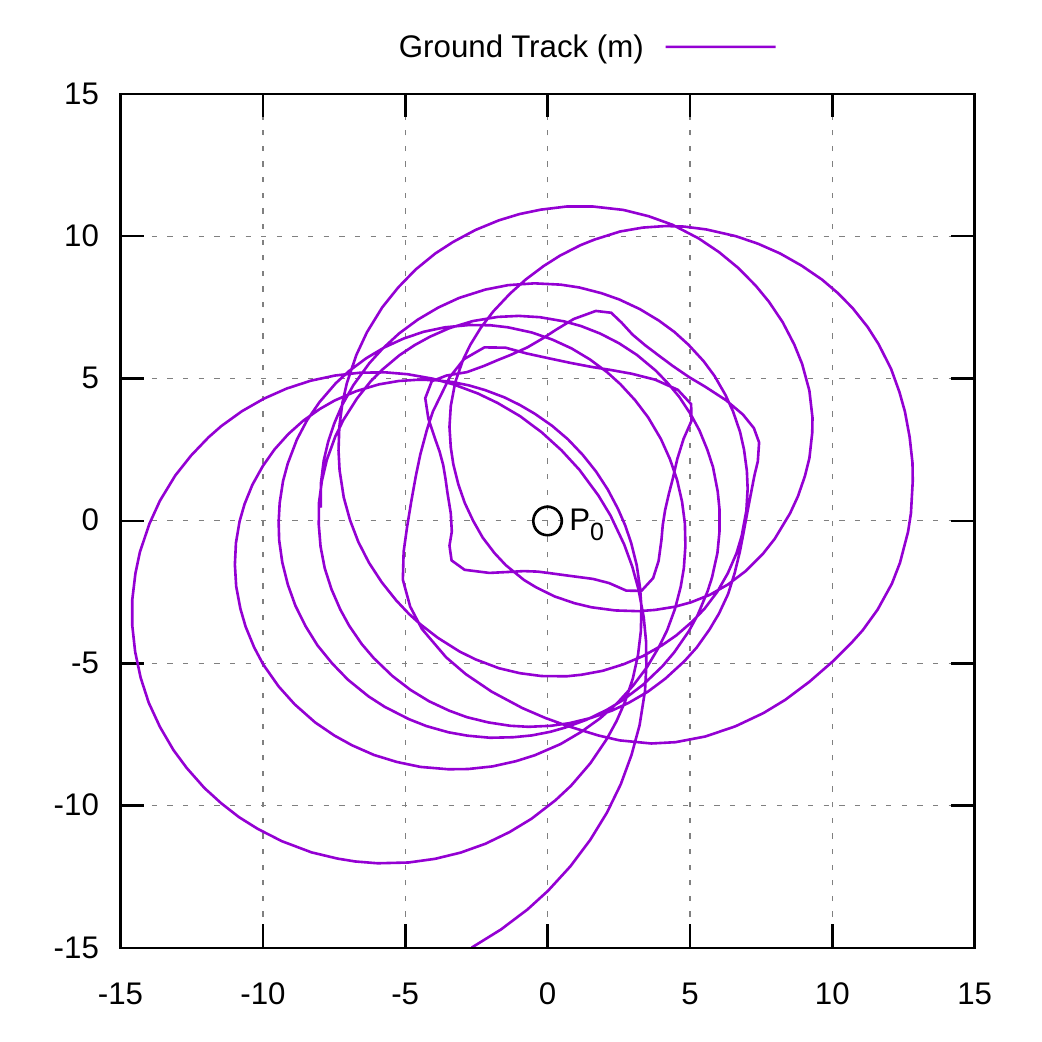}\includegraphics[width=0.49\columnwidth]{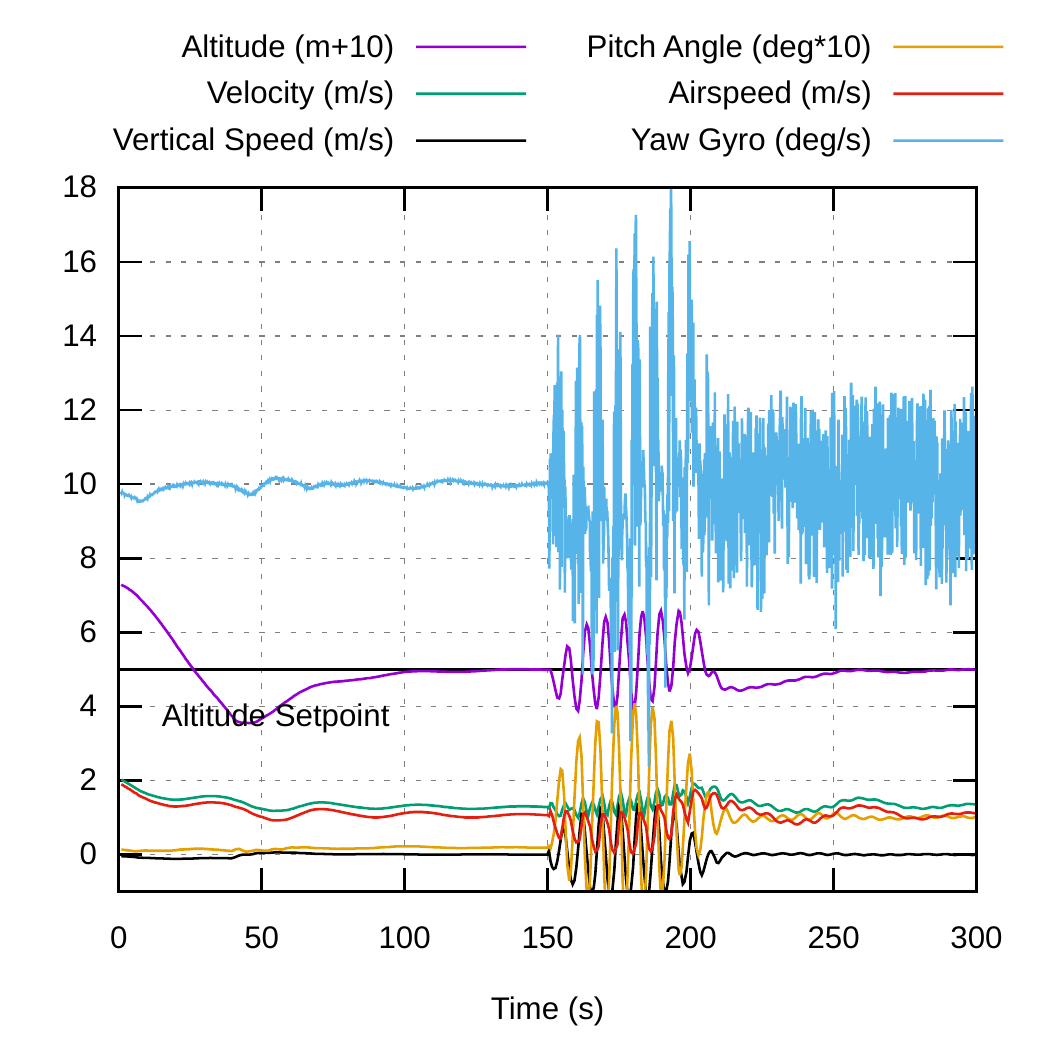}}\\
 {\small{}{}Trajectory Following:}\\
 {\small{}{}\includegraphics[width=0.49\columnwidth]{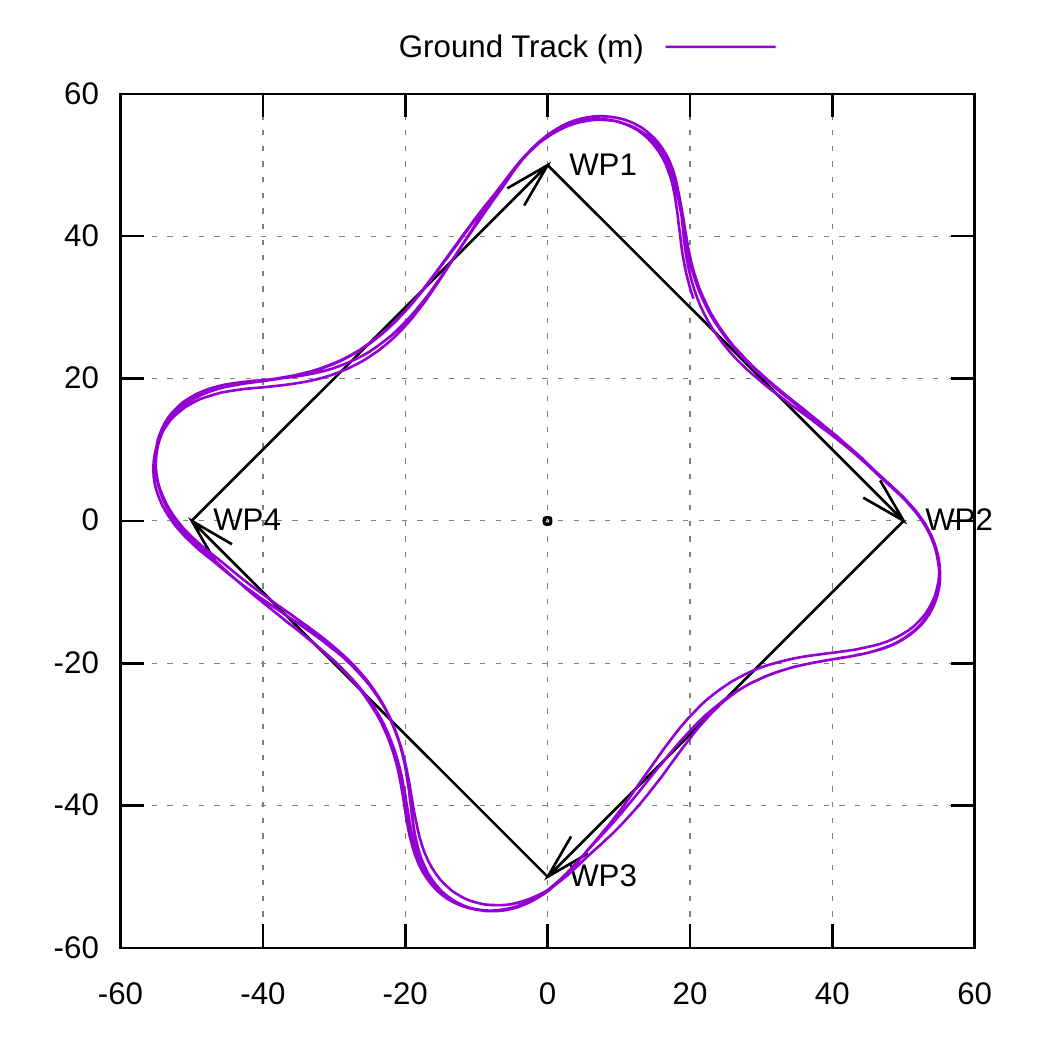}\includegraphics[width=0.49\columnwidth]{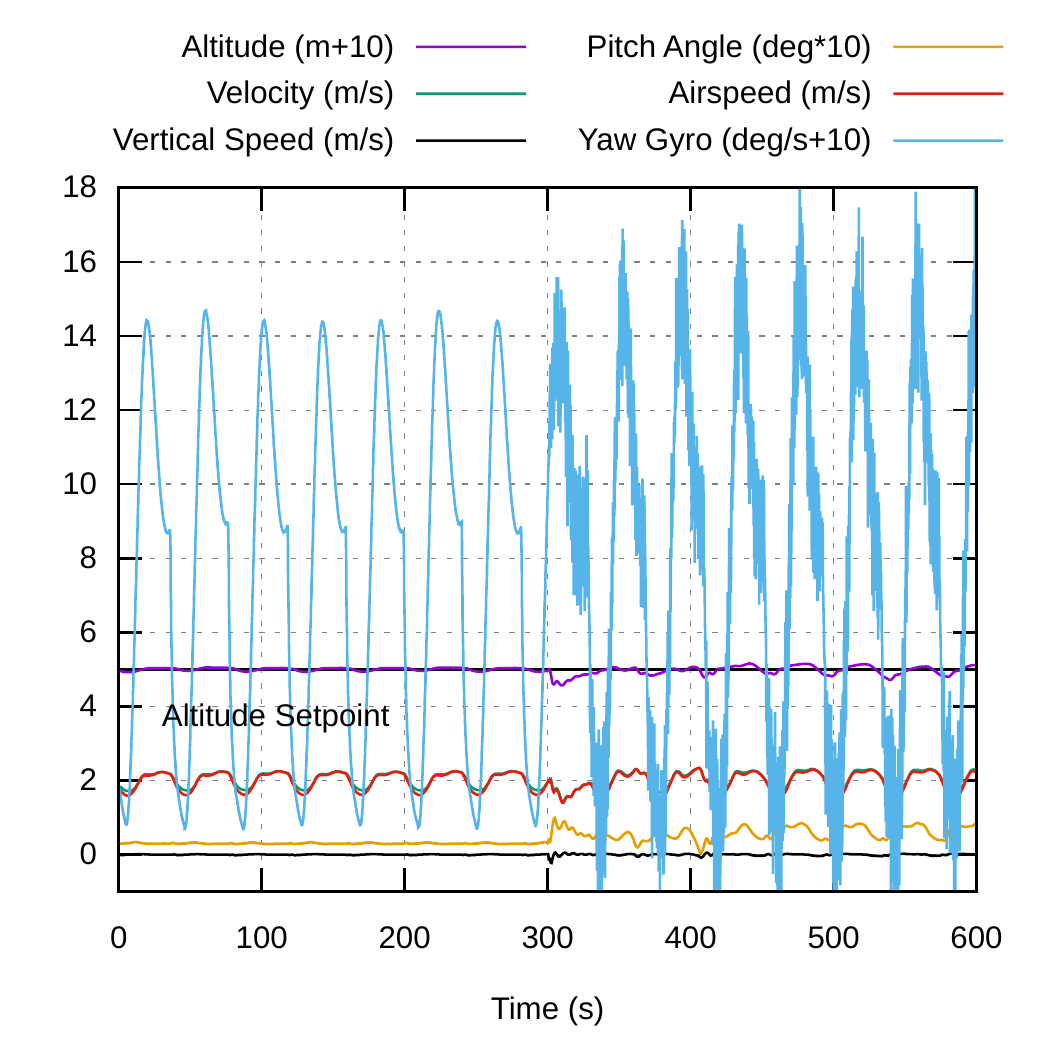}}{\small\par}
\par\end{centering}
\begin{centering}
 
\par\end{centering}
{\small{}{}\caption{{\small{}{}\label{fig:Sim exp 5}}\textbf{\small{}{}Simulation experiment 5}{\small{}{} - The simulated blimp was partially deflated at the half time mark of each experiment (``}\textsf{\small{}{}deflate\_blimp.sh /blimp 8 0 0.2 .95}{\small{}{}''). In the loiter experiment, this exposes a severe oscillation instability in altitude and pitch that was not present when fully inflated, although the blimp is still able to control itself. In both situations, the altitude accuracy is noticeably decreased. In this plot, the yaw gyroscope signal is plotted, revealing a high frequency oscillation in yaw, as the actuation of the rudder now moves the fin that the simulated gyroscope is mounted to, relative to the blimp.}}
}{\small\par}

\end{figure}

\noindent - We re-conducted both simulation experiment 2 and simulation experiment 3. During the experiment we adjusted the rigidity of the simulated blimp to simulate helium loss and observe the effects. Simulated loss of hull pressure and the resulting decrease in rigidity reduced the controllability of the blimp. A reduction in altitude accuracy and oscillations in both pitch, altitude and also in course were observed. The ability of the fins to flex and twist in response to control actuation not only adds additional oscillation modes, especially for the top fin which houses the IMU (both in simulation and on the real-world blimp), it also reduces the control effects, since the fins rotate in response to the control forces until a new force equilibrium is reached. This significantly reduces the overall control force enacted on the hull, since the angle of attack of the fin and attached rudder negate each other. Nevertheless, the blimp remains controllable, as long as sufficient buoyancy is present to maintain altitude (Fig. \ref{fig:Sim exp 5}). 

\subsection{Real-world experiment}

\subsubsection{Loitering outdoor with wind and deflation}

\begin{figure}
\noindent \begin{centering}
\includegraphics[width=0.49\columnwidth]{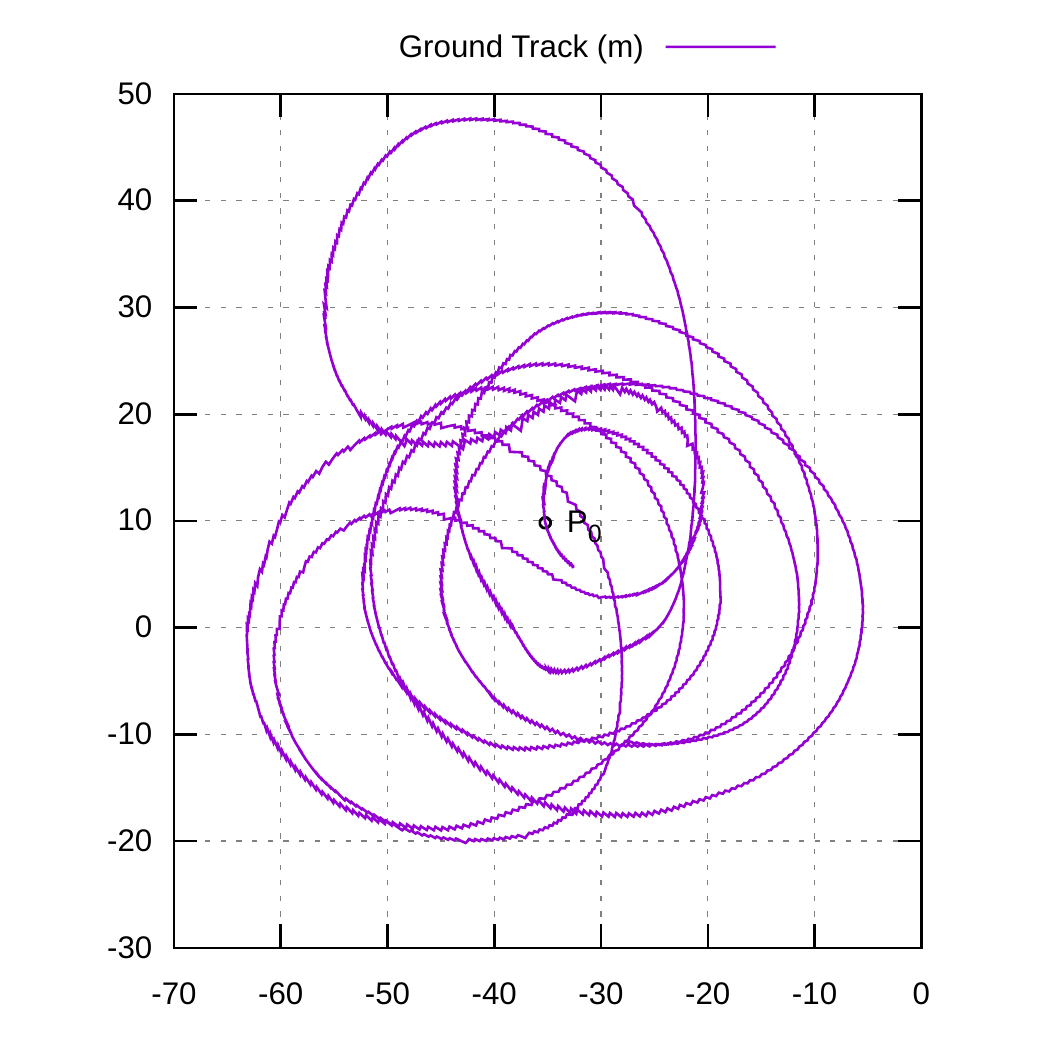}\includegraphics[width=0.49\columnwidth]{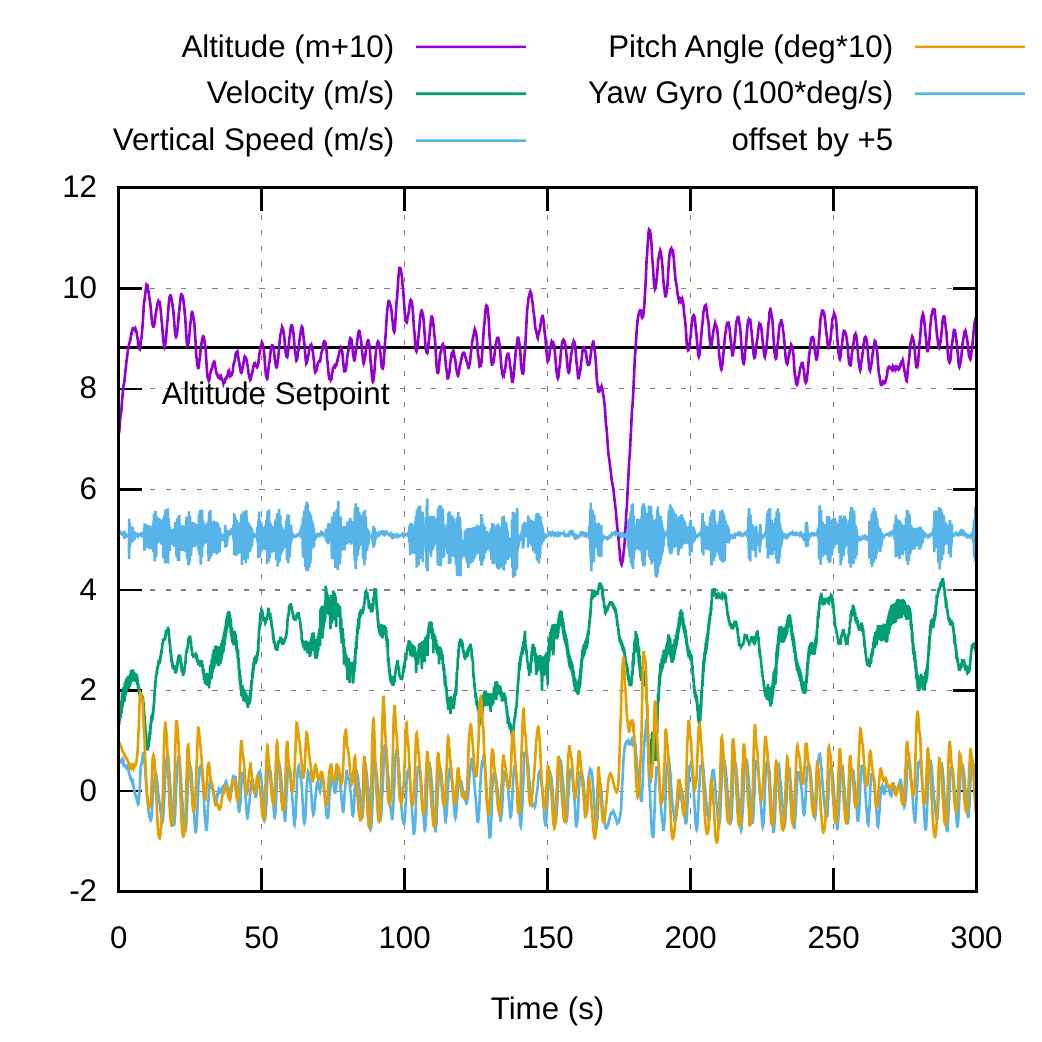} 
\par\end{centering}
\caption{{\small{}{}\label{fig:Real exp 2}}\textbf{\small{}{}Real-world experiment}{\small{}{} - The blimp loiters around a given position in the physical world, while losing helium. The winds were benign, but a single gust at the $160\mbox{s}$ mark results in a noticeable drop in altitude which the control subsequently corrected, similar to those seen in simulation experiment 4. Due to lack of functional airspeed sensor and higher than tuned-for thrust, the ground speed and turn radius are larger than in the simulation. As predicted in simulation experiment 5, the blimp displays pitch-altitude oscillations throughout the flight. A high frequency oscillation in the yaw axis was also present (drawn down-scaled by factor $100$), as predicted.}}
\end{figure}

\noindent - We configured the real-world blimp with the control coefficients determined in simulation and engaged position hold mode. Due to a previous ground test incident, the hull had developed significant leakage. As such we were able to reproduce the effects in simulation experiment 5 in real-world, with the blimp's internal pressure far below optimum and dropping. Although we were able to maintain buoyancy by removing ballast weights, the blimp's flight performance was below optimal. As predicted in simulation experiment 5, a high frequency oscillation developed in the yaw axis. The rudder actuation imparted momentum and flexion on the top fin which was picked up by the IMU gyroscope. The real-world blimp was flown with identical control coefficients to those used in simulation experiments. However, due to changes on the electronic speed controller (ESC), the blimp produced higher thrust than anticipated. This might have contributed to oscillations in pitch and altitude, which were similar, but more persistent and of higher frequency than those observed in simulation experiment 5. Our airspeed sensor failed to work and had to be deactivated, leading to $v_{I}$ estimated based on ground speed, but we expect minor impact due to the benign winds that day ($<1\mbox{m/s}$ average). The altitude and position were controlled reliably for more than $5$ minutes of continuous flight time. One observed wind gust caused an altitude excursion, as predicted by simulation experiment 4. No intervention by the pilot was required at any time (Fig. \ref{fig:Real exp 2})\footnote{Video, real-world experiment: \href{https://youtu.be/KF4Dni7-Jnw}{https://youtu.be/KF4Dni7-Jnw}}. 

\subsection{Discussion}

Our experiments have confirmed, that changes in rigidity, as simulated in our framework, have significant effects on behavior and control of airships, which is well documented by the literature \cite{LI2011217}. Modeling of wind gusts has also exposed notable control behavior, not seen in non-turbulent air. We have shown, that a simulation framework as presented here can be used to develop, tune and evaluate a control algorithm with results comparable to and representative of the real world, with limitations discussed below.

%% file: conclusion.tex
\section{Limitations\label{sec:Limitations}}

Very large airships would benefit from wind and turbulence simulation, that accurately models local variations of the wind flow at the same point in time. This would require the wind model to be modified, to estimate a whole wind field and calculate the wind vector as a function of both space and time for multiple places. Due to the modular approach taken with aerodynamics, it would be relatively straightforward to incorporate this data into a simulation with a sufficiently segmented hull, although this is beyond the scope of our current work.

Our simulation, as conducted for our experiments, is not quantitatively accurate for our vehicle. To achieve quantitative accuracy, or to analyze the simulation error made by the model, it would be necessary to compare the calculated forces to those determined in wind tunnel tests of CFD analysis, which could be done with acceptable effort for the rigid case. This data could be used to solve for ``best fitting'' coefficients on all component objects, to minimize the simulation error. This would also show, if additional correction coefficients are needed to model aerodynamic interaction between different components, or if the current set of coefficients in combination with higher granularity is sufficient. We would defer this analysis to future work.

Our current aerodynamic model can not model ``prop wash'' and the effects it has on impinged control surfaces and fins. This should be addressed by future work to improve the blimp simulation, although solving this would also be beneficial for other vehicles such as multicopters and fixed wing.

Although our model can simulate deformations of the hull as a whole, it can not model surface deformations of the skin such as wrinkling or denting. Temporary changes of the geometric shape of a component could have significant effects on the drag coefficient as mentioned in \cite{LI2011217}. These have been observed to occur in our real-world experiment. Although it would be straightforward to adjust the aerodynamic coefficients of component objects in the simulation, this would require knowledge of the relationship between aerodynamic forces, internal pressure, shape and effects on aerodynamics. This is a hard problem and has not been well researched yet, as noted in \cite{LI2011217}. It justifies future work.

\section{Summary\label{sec:Summary}}

We have created a solution for SITL and HITL simulation as well as a control algorithm, which allows researchers to set up and fly LTAVs with the same tools and flight electronics that have been used for years with fixed wing and multicopters. The seamless integration with ROS and Gazebo incorporates airship support into popular robotic tool chains. As such, this work should facilitate a significant reduction of the gap between heavier- and lighter-than-air UAV robotics. Although LTAVs can be cumbersome to work with due to size and lifting gas handling, turning them into autonomous research platforms is now straightforward. Researchers can attach a COTS flight controller and run freely available open-source tools on any flight hardware. This had previously been possible only for heavier-than-air craft.

We have shown that modeling both wind turbulence and deformation is crucial for realistic simulation of LTAVs. Rigid models in non-turbulent air can not predict important aspects of the vehicle's behavior.

We presented a reproducible approach for modeling non-rigidity in simulation, in combination with wind turbulence. This allows researchers to test, evaluate, or, in case of learning algorithms, to ``train'' control algorithms against these conditions. The simulation environment is well suited for multi vehicle experiments and allows research on mixed formations involving both ground robots, heavier-than-air and LTAV. Although our current simulation models a specific blimp, it is straightforward to modify the configuration for other LTAVs. Only the robot XML/URDF configuration has to be changed.

The authors consider this work to be the foundation for their own future lighter-than-air research and hope the scientific community will find it useful as well.

%% file: bibliography.tex
% ---- Bibliography ----

 \bibliographystyle{llncs2e/splncs}
\phantomsection\addcontentsline{toc}{section}{\refname}\bibliography{paper}